\documentclass[sigplan,10pt,nonacm]{acmart}
\usepackage{array}
\usepackage{makecell}
\usepackage{booktabs}
\usepackage{colortbl}
\usepackage{multirow}
\usepackage{algorithm}
\usepackage{algpseudocode}
\usepackage{subcaption}
\usepackage{enumitem,xspace}
\renewcommand\footnotetextcopyrightpermission[1]{}
\AtBeginDocument{%
  }

\setcopyright{acmlicensed}
\copyrightyear{2018}
\acmYear{2018}
\acmDOI{XXXXXXX.XXXXXXX}
\acmConference[Conference acronym 'XX]{Make sure to enter the correct
  conference title from your rights confirmation email}{June 03--05,
  2018}{Woodstock, NY}
\acmISBN{978-1-4503-XXXX-X/2018/06}
\usepackage[font=small]{caption}

\newcommand{\sys}{Libra\xspace}

\begin{document}

\title[]{\sys: Efficient Resource Management for Agentic RL Post-Training}


\author{
    Kaiwen Chen\textsuperscript{\rm 1},
    Xin Tan\textsuperscript{\rm 1},
    Jingzong Li\textsuperscript{\rm 2},
    Zhi Zhou\textsuperscript{\rm 3},
    Cen Li\textsuperscript{\rm 3},
    Jiang Liu\textsuperscript{\rm 3},
    Jie Meng\textsuperscript{\rm 3},
    Jiazhi Jiang\textsuperscript{\rm 1},\\
    Hong Xu\textsuperscript{\rm 1}
    \\
    \textsuperscript{\rm 1}The Chinese University of Hong Kong\\
    \textsuperscript{\rm 2}The Hang Seng University of Hong Kong\\
    \textsuperscript{\rm 3}Independent Researcher\\
}


\begin{abstract}
Reinforcement learning (RL) has emerged as a standard post-training paradigm for shaping large language models (LLMs) into capable agents. 
In agentic RL, the rollout stage generates trajectories while invoking tools, producing long-tailed and non-stationary workloads that expose two fundamental challenges.
First, due to the long-tailed response distribution, a small fraction of trajectories dominates rollout makespan.
Second, rollout and training differ in their compute patterns, memory demands, and sensitivity to sequence length. As the policy evolves, shifts in the workload distribution further change their relative resource demands, making it difficult to maintain balanced execution across the two stages.

We present \sys, an adaptive runtime for agentic RL post-training with two complementary components:
(1) intra-stage scheduling via a Causality-Guided Bucket Scheduler that routes requests across execution buckets with different parallelism configurations, reducing delays from rollout stragglers; and
(2) cross-stage coordination that dynamically reallocates workers between rollout and training as the workload changes. It moves workers between the two stages through a non-blocking protocol without interrupting ongoing training.
Evaluated on a 48$\times$ NVIDIA A800 GPU cluster and a 160$\times$ Ascend 910B3 NPU cluster across three agentic benchmarks, \sys achieves up to 4.2$\times$ higher throughput and up to 2.7$\times$ faster reward convergence. 
\end{abstract}




\settopmatter{printfolios=true}

\maketitle
\thispagestyle{plain}
\fancyhead{}
\section{Introduction}

Reinforcement learning (RL) has become an important post-training paradigm for large language models (LLMs), extending from alignment with human preferences~\cite{RLHF1,RLHF2,rlhf3} and complex reasoning~\cite{deepseek-r1,kimi-k1.5,llm-posttraining-deepdive} to the development of agentic capabilities~\cite{agentic_rl_1,verltool,retool,tora,react,gemini,agenticRL_survey}.
A typical agentic RL iteration consists of three stages: trajectory generation (rollout), trajectory evaluation, and policy update (training).
During rollout, the policy model generates trajectories through autoregressive decoding and interactions with external tools.
During training, the collected trajectories and their rewards are used to compute advantages and update the policy.
Trajectory evaluation assigns rewards to generated trajectories and is comparatively lightweight~\cite{deepseek-r1,swe-bench,ppo}, leaving rollout and training as the dominant computational stages~\cite{Laminar,areal,streamrl}.
Overall pipeline throughput therefore depends primarily on how quickly rollout produces trajectories and training consumes them.

During agentic RL post-training, the model invokes external tools and incorporates their outputs into subsequent generation.
The resulting workload has three important properties.
First, trajectories are generated online so their lengths are determined at runtime.
Second, tool invocations cause sequence lengths to fluctuate widely: sequence length varies substantially with tool execution outcomes, such as payload size and return status.
Third, the sequence length distribution shifts continuously as the policy improves, making the workload non-stationary. 

Efficiently executing this workload requires addressing two challenges.
The first is mitigating the impact of long-tailed trajectories on rollout performance.
Although most trajectories are short, a small fraction are substantially longer and can dominate the rollout makespan.
Existing systems largely attempt to mitigate the long-tailed problem through upfront length prediction at $T=0$ to classify requests and apply differentiated execution strategies.
However, their prediction mechanisms become less reliable in agentic RL.
Methods that exploit intra-group or cross-epoch response similarity~\cite{Seer,RhymeRL} assume that sequence length is largely prompt-determined, an assumption that breaks down in agentic RL where mid-trajectory tool execution outcomes heavily alter sequence length.
Other approaches deploy auxiliary pre-trained language models or embedding-based predictors~\cite{Heddle,streamrl}; however, as the actor policy continuously updates during RL post-training, these parametric predictors suffer severe representation drift and require frequent, compute-intensive retraining on GPUs.

The second is the cross-stage coordination challenge.
Rollout and training act as networked peers that continuously exchange state: training propagates updated weights to rollout, while rollout streams generated trajectories back to training. 
However, they differ fundamentally in compute patterns, memory demands, and sensitivity to sequence length. Rollout is memory- and bandwidth-bound and scales linearly with sequence length,
whereas training is compute-bound and amortizes length variation through batching.
Compounding the problem, the sequence length distribution drifts continuously as the policy evolves, rendering any static resource allocation progressively suboptimal.
Existing RL frameworks either optimize rollout in isolation under static stage partitions~\cite{verl,areal,Laminar} or merely reallocate GPU counts across rigid parallelism boundaries~\cite{DynaRL}, failing to align resources with sequence-length demands. 
Meanwhile, distributed parallelization frameworks~\cite{alpa,galvatron,sailor} target LLM pre-training, which are not designed to model the closed-loop state coupling, asymmetric scaling, and non-stationary drift inherent to RL post-training.

To address these challenges, we present \sys, an adaptive runtime designed for the disaggregated, asynchronous pipeline of agentic RL post-training. 
\sys treats rollout and training as co-designed, networked peers, establishing adaptive execution across two complementary levels:
(1) intra-stage scheduling that performs causality-driven request routing across heterogeneous TP buckets, and 
(2) cross-stage coordination that coordinates resources and elastically re-balances workers across rollout and training over high-speed interconnects.

\noindent \textbf{Intra-stage scheduling:} 
To eliminate rollout stragglers, our first insight is that long-tailed problem can be effectively mitigated by matching requests to differentiated Tensor Parallelism (TP) configurations.
A small TP degree minimizes cross-GPU communication overhead and maximizes token throughput for short sequences, whereas a large TP degree alleviates KV-cache memory pressure and sustains high decode efficiency for long sequences.
Partitioning the rollout cluster into heterogeneous execution buckets with varying TP degrees (e.g., TP-1 to TP-8) allows requests of different sequence lengths to run under their optimal parallelism configurations, 
substantially reducing the straggler effect of long-tailed trajectories.
However, dynamically steering requests to matching TP buckets requires timely length estimation.
Here, \sys makes a second key observation: in agentic RL, tool execution outcomes serve as runtime signals that are informative of remaining sequence length.
For example, a failed tool call triggers diagnostic retry chains that predictably inflate sequence length, while a massive tool return payload deterministically consumes substantial context. 
Building on these insights, \sys introduces the Causality-Guided Bucket Scheduler (\S\ref{CGBS}).
The scheduler tracks execution history using a lightweight causality-aware prefix tree (trie). 
Each node is keyed by the observed sequence of composite tool states (tool types, payload size categories, and success/failure statuses) and stores an empirical histogram of residual sequence lengths. 
Upon each tool return, the scheduler performs an $O(1)$ trie lookup to retrieve the residual length distribution and casts routing as an online cost-benefit decision.

\noindent \textbf{Cross-stage coordination:} 
At the cluster level, \sys dynamically maintains balanced execution across rollout and training under continuous workload drift (\S\ref{sec:resource_planner}). 
Under a fixed cluster GPU budget, a global resource planner jointly optimizes both the GPU allocation across stages and the parallelism strategy within each stage. 
Guided by an analytical Cost Evaluator (CE), the planner explores the multi-dimensional training strategy space ($(TP, EP, PP, DP)$) via a topology-aware pruned decision tree, while determining optimal rollout heterogeneous bucket capacities through a dynamic programming formulation.

To enact these reallocations without interrupting ongoing execution, \sys organizes the cluster into three pools: a Core Training Pool, a Core Rollout Pool, and an Elastic Hybrid Pool. 
\sys recognizes that elasticity need not trigger global communicator rebuilds: by enforcing that elastic resources enter or exit strictly at the granularity of complete, self-contained Data-Parallel (DP) replicas, the core training topology remains immutable. 
\sys decouples communication into permanent intra-replica collective domains and a dynamic inter-replica RDMA control plane. 
Through a non-blocking joining protocol, hybrid DP replicas transition from rollout to training. They fetch model and optimizer snapshots asynchronously via RDMA during active training forward/backward steps, aligning optimizer moments at step boundaries without stalling ongoing training.
Transient imbalances are absorbed immediately by hybrid workers toggling between rollout inference and training DP replicas, while persistent distribution shifts trigger planned repartitioning of core pools during inherent pipeline slack windows.

We evaluate \sys on both a 48-GPU cluster of NVIDIA A800s and a 160-NPU cluster of Ascend 910B3s across three representative agentic RL benchmarks: Search-R1~\cite{searchr1}, R2E-Gym~\cite{r2egym}, and DAPO-Math-17K~\cite{DAPO}. 

\sys consistently outperforms state-of-the-art baselines, achieving up to 4.2$\times$ higher pipeline throughput and 2.7$\times$ faster reward convergence. \sys has been open-sourced on https://github.com/NetX-lab/Libra.

In summary, this paper makes three main contributions:
\begin{itemize}[nosep,leftmargin=*]
\item \textbf{Intra-stage Causality-Guided Scheduling}: We propose the Causality-Guided Bucket Scheduler, which leverages mid-trajectory tool returns as runtime signals within a prefix-tree structure to predict residual sequence lengths, routing requests across heterogeneous TP buckets. 
\item \textbf{Cross-stage Joint Planning and Elastic Coordination}: We design a global planner that jointly optimizes cross-stage resource splits and intra-stage parallelism strategies, coupled with an elastic execution mechanism that rebalances workers between rollout and training via decoupled communicators and a non-blocking joining protocol. 
\item \textbf{Implementation and Comprehensive Evaluation}: We implement \sys and evaluate it on 48$\times$ NVIDIA A800 GPUs and 160$\times$ Ascend 910B3 NPUs, demonstrating up to 4.2$\times$ throughput speedup and 2.7$\times$ faster reward convergence across diverse agentic workloads. 
\end{itemize}

\section{Background and Motivation}
\label{sec:motivation}
We start by presenting the background and motivation.

\subsection{Disaggregated and Asynchronous RL}
Recent RL frameworks commonly execute rollout and training in a disaggregated and asynchronous manner~\cite{areal,roll,Distrl,LlamaRL} to better accommodate the distinct execution characteristics of each stage.
The memory- and bandwidth-bound rollout runs on a separate cluster of GPU workers, which continuously generate trajectories with slightly stale weights, while the compute-intensive training cluster periodically pushes updated parameters back to the rollout side~\cite{areal,Laminar,streamrl,RollArt,ROSE,Distrl,asyncflow}.

\begin{table}[t]
    \centering
    \begin{minipage}[t]{0.42\columnwidth}
        \centering
        \vspace{2.1mm}
        \small
        \setlength{\tabcolsep}{1pt}
        \resizebox{\linewidth}{!}{
            \begin{tabular}{@{}l r r r r r l@{}}
            \toprule
            \makecell[c]{Case} & \makecell[c]{Tool-call\\Count} & \makecell[c]{Tool\\Tokens} & \makecell[c]{Failure\\Num} & \makecell[c]{Total\\Tokens} & \makecell[c]{Label} \\
            \midrule
            \makecell[c]{django\\14787} &  \makecell[c]{8} & \makecell[c]{1,029} & \makecell[c]{0} &  \makecell[c]{7,363} & \makecell[c]{Small\\Payload} \\
            \midrule
            \makecell[c]{scikit-learn\\11310} &  \makecell[c]{9} & \makecell[c]{12,461} & \makecell[c]{0} & \makecell[c]{21,220} & \makecell[c]{Large\\Payload} \\
            \midrule
            \makecell[c]{django\\10999} &  \makecell[c]{46} & \makecell[c]{3,962} & \makecell[c]{3} &  \makecell[c]{40,960} & \makecell[c]{Failure\\Cascade} \\
            \midrule
            \makecell[c]{django\\16560} & \makecell[c]{50} & \makecell[c]{7,809} & \makecell[c]{8} & \makecell[c]{40,960} & \makecell[c]{Failure\\Cascade} \\
            \bottomrule
            \end{tabular}
        }
        \caption{Trajectory divergence across tasks in R2E-Gym \cite{r2egym}, generated by Qwen3-14B.}
        \label{tab:trajectory_divergence}
    \end{minipage}
    \hfill
    \begin{minipage}[t]{0.54\columnwidth}
        \centering
        \vspace{4pt}
        \includegraphics[width=\linewidth]{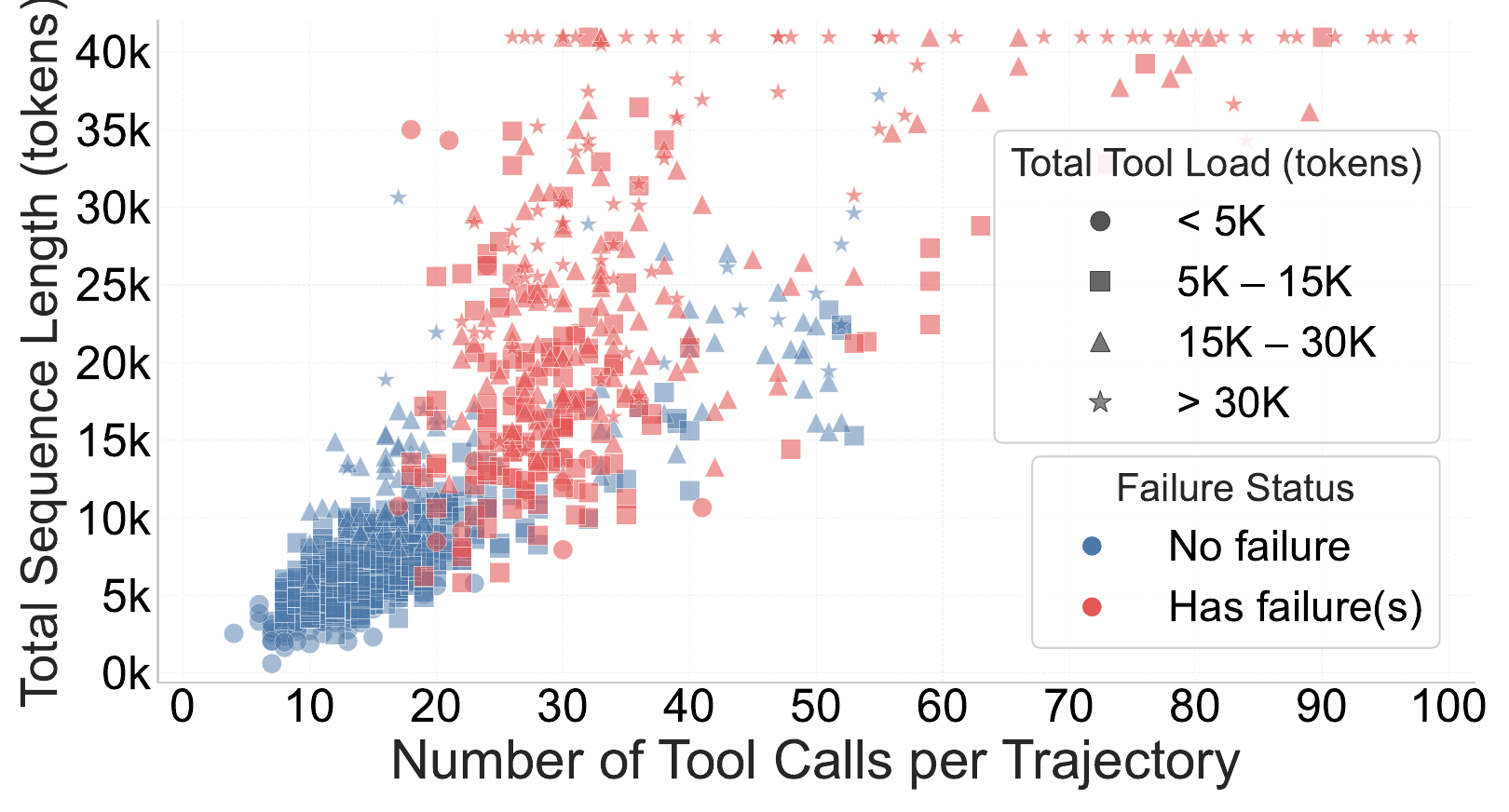}
        \vspace{-7mm}
        \captionof{figure}{Tool-call count versus sequence length in R2E-Gym \cite{r2egym}, generated by Qwen3-14B.}
        \label{fig:tool_calls_vs_length}
    \end{minipage}
\end{table}

\subsection{Characteristics of Agentic RL Post-Training}
Agentic RL post-training workloads exhibit four properties:

\noindent\textbf{Long-tailed Rollout.} 
A large body of work~\cite{Laminar,Seer,streamrl,Heddle,stage_fusion,roll_packer} has identified the long-tailed output sequence as a major source of rollout inefficiency.
Although most trajectories conclude within a short reasoning process, a small fraction is substantially longer. 
These long-tailed sequences dominate rollout makespan, forcing faster-finishing workers to idle while waiting for the slowest trajectories to complete. 
According to our measurements on R2E-Gym \cite{r2egym} (Qwen3-14B), the longest 10\% trajectories account for over 50\% of total rollout time.

\noindent \textbf{Causal Drivers of Sequence Length.}
To our knowledge, we are the first to identify and systematically characterize this phenomenon: 
In agentic RL, sequence length is causally driven by external tool invocations at runtime.
Tool-return payload size and success/failure status are the principal causal drivers of trajectory expansion, introducing substantial runtime variability beyond the initial prompt.
Table~\ref{tab:trajectory_divergence} demonstrates this causal relationship across representative tasks in R2E-Gym~\cite{r2egym}.
First, tool-return payload size directly dictates context expansion: a lightweight payload (django-14787, 1K tool tokens) yields a concise trajectory of $<8$K tokens, whereas an extensive tool return (\texttt{scikit-learn-11310}, 12.5K tool tokens) abruptly inflates the sequence to over 21K tokens.
Second, tool execution status acts as an even more decisive causal trigger: when tool calls fail, the model enters diagnostic retry loops and error-correction steps.
As shown in \texttt{django-10999} and \texttt{django-16560}, such failure cascades multiply tool invocations by over $5\times$, driving total length to the 41K context ceiling.
Figure~\ref{fig:tool_calls_vs_length} corroborates this causal link across the full benchmark: failed trajectories systematically concentrate in the high-call, extreme-length regime, while successful trajectories remain in lower-length tiers.

\begin{figure}[t]
    \centering
    \includegraphics[width=\columnwidth]{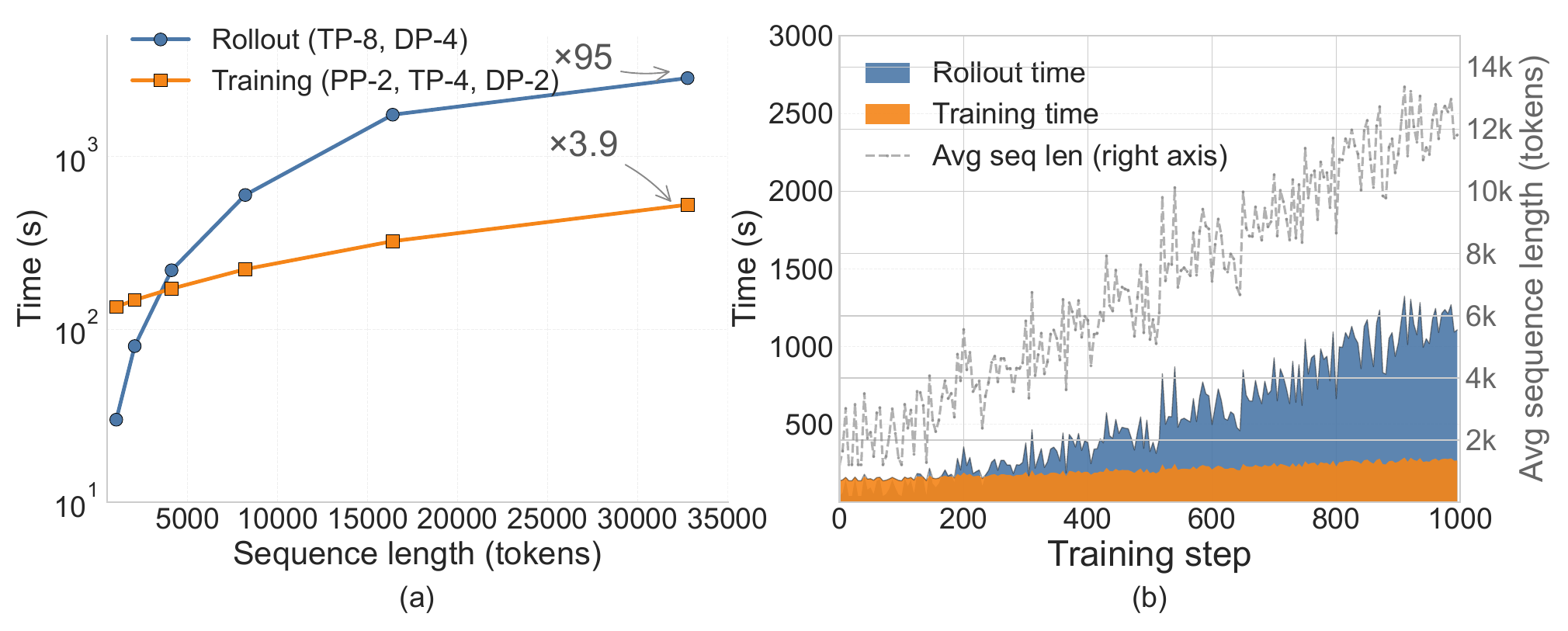}
    \caption{
    (a)~Comparison of average latency between rollout and training as output sequence length increases.
    (b)~Workload drift over the course of training.
    We use Qwen3-32B-Base on A800 80GB GPUs; the rollout stage runs on 32~GPUs with TP-8 and DP-4 (batch size~512), and the training stage runs on 16~GPUs with PP-2, TP-4, and DP-2 (global batch size~4096, mini-batch size~16) on AIME~\cite{aime}.
    }
    \label{fig:stage_asymmetry}
\end{figure}

\noindent \textbf{Non-stationary Nature of the Workload.}
As RL post-training proceeds, the actor model's reasoning capability is continuously updated, which alters its response patterns gradually.
Figure~\ref{fig:stage_asymmetry}(b) illustrates one manifestation of this drift: on our agentic RL benchmark, the average sequence length grows from $\sim$2,500 to $\sim$11,500 tokens, causing rollout time to gradually overtake training time.
We note that the direction and magnitude of length drift are not fixed---they depend on the model architecture, the RL algorithm, and the task domain (e.g., some workloads may exhibit length contraction rather than expansion~\cite{contraction1,kimi-k1.5,enough_thinking}).
What is guaranteed, however, is that the sequence length distribution shifts as the policy evolves, so any statically chosen resource split progressively diverges from the optimal allocation.

\noindent \textbf{Strong Asymmetry Between Rollout and Training.}
Figure~\ref{fig:stage_asymmetry}(a) shows that when sequence length grows from 1K to 32K tokens, rollout latency increases by 95$\times$ (from 30~s to 2,850~s), whereas training time increases by only 3.9$\times$ (from 135~s to 527~s).
Rollout latency grows tremendously with sequence length due to autoregressive decoding, while training amortizes length variation through batching.

\subsection{Opportunities}
These characteristics motivate us to optimize agentic RL post-training along two complementary axes.

\begin{figure}[t]
    \centering
    \includegraphics[width=\columnwidth]{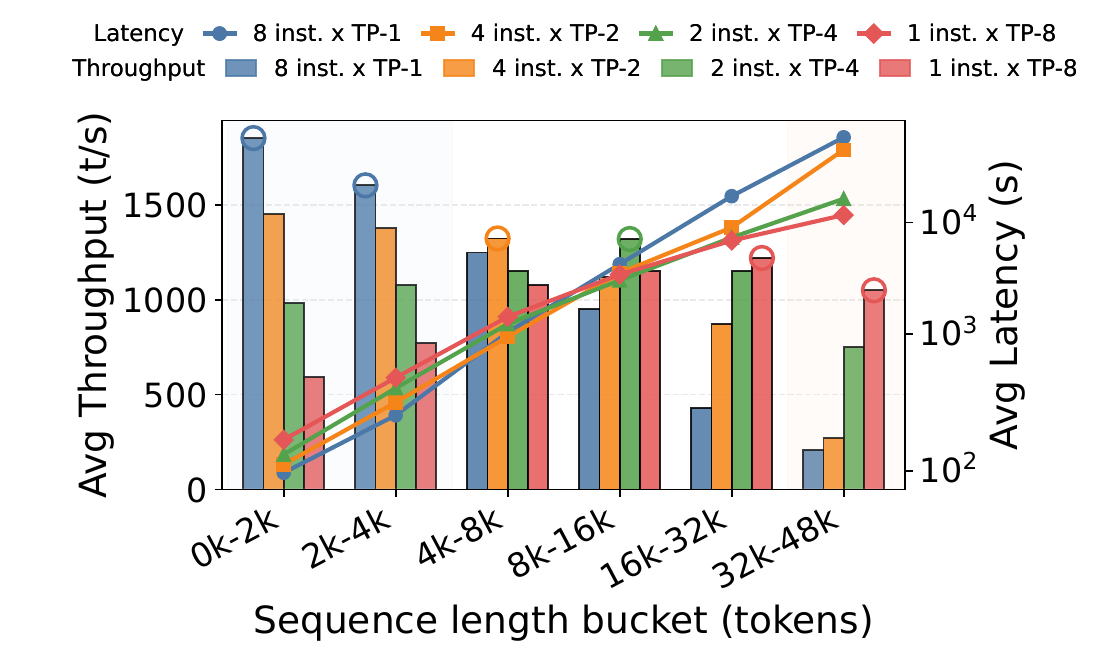}
    \caption{Average throughput vs. sequence length across TP configurations on 8xA800 GPUs (Qwen3-14B, batch size 512, R2E-Gym). ``2 inst. x TP-4'' denotes two instances each using 4-GPU Tensor Parallelism.}
    \label{fig:heterogeneity}
\end{figure}

\label{section:opportunity}
\noindent \textbf{Rollout Optimizations.} 
Rollout efficiency in agentic RL can be improved by two ideas. 
First, the sharp divergence in sequence length motivates us to consider heterogeneous parallelism configurations, especially Tensor Parallelism (TP), tailored to different sequence lengths. 
Figure~\ref{fig:heterogeneity} shows that TP-1 achieves 1,852~token/s for short sequences (0k--2k) but collapses to 430~token/s at 16k--32k due to KV-cache pressure; conversely, TP-8 starts at only 591~token/s for short sequences but sustains 1,220~token/s at 16k--32k, outperforming TP-1 by $2.8\times$  in our experiment.
Large TP partitions the KV cache and model weights across more GPUs, mitigates memory bottlenecks, and amortizes the all-reduce overhead over heavier computation for long sequence; small TP reduces communication overheads, matching the low memory needs profile of short requests.
Thus, rather than adopting a uniform TP configuration across the rollout cluster, we concurrently deploy multiple TP configurations within the cluster.

The cluster is partitioned into buckets, where each bucket hosts inference instances with a distinct TP degree optimized for requests within certain ranges of sequence lengths.
We focus on TP because Pipeline Parallelism (PP) can exacerbate straggler effects through pipeline bubbles under variable-length trajectories, 
and Expert Parallelism (EP) addresses expert load balancing rather than sequence-length-induced stragglers. 
Further details are discussed in Appendix~\ref{appendix:pp_ep}.

The next natural idea is request-bucket assignment based on runtime causal signals.
A straightforward approach is to predict each trajectory's final length and dispatch it to the matching bucket using a dedicated model~\cite{embedding,Heddle}.
However, as the policy drifts during the course of training, the prediction accuracy will be degraded without periodical retraining.
An alternative is prediction-free heuristics like the classical MLFQ, which promotes requests to larger-TP buckets only after their lengths cross fixed thresholds.
It is both delayed and incremental: trajectories suffer latency at ill-matched configurations before any threshold is crossed, and long trajectories endure multiple migrations with each incurring state-transfer overhead before reaching their final bucket.

The Causality-Guided Bucket Scheduler exploits the causal signals from the tool-calling results to make routing decisions.
A trajectory encountering a heavy payload or a tool-execution failure can be promoted to a larger TP bucket because its sequence length is very likely to expand, while one with lightweight tool returns may remain on its current setup.
This assignment is dynamic during the request's lifetime and resolves the limitations of both baselines.
First, \sys naturally adapts to policy drift: when RL optimization shifts the trajectory distribution, \sys only needs to update the decision tree offline instead of retraining a prediction model.
Second, \sys enables earlier and decisive migration: whereas MLFQ upgrades a request only after its length crosses a threshold, our bucket scheduler makes a routing decision immediately upon each tool return, often migrating a request to its final bucket in a single step rather than through gradual promotion.

\noindent \textbf{Cross-Stage Optimizations.} 
Since rollout and training execute concurrently, the end-to-end iteration time is dictated by whichever stage runs slower under the current GPU allocation. The bottleneck is a dynamic consequence, not a fixed property of rollout. 
This exposes a clear opportunity: by dynamically reallocating GPUs from the faster stage to the slower one, the system can mitigate the bottleneck and reduce the overall iteration time. 
However, the optimal allocation is not static, as the workload is non-stationary and the balance point drifts over time. 
A system that continuously tracks this drift and uses lightweight elasticity to reassign GPUs without disrupting execution can sustain higher throughput than any static allocation.

\section{\sys Overview}
\label{sec:overview}

\begin{figure}[t]
    \centering
        \includegraphics[width=\columnwidth]{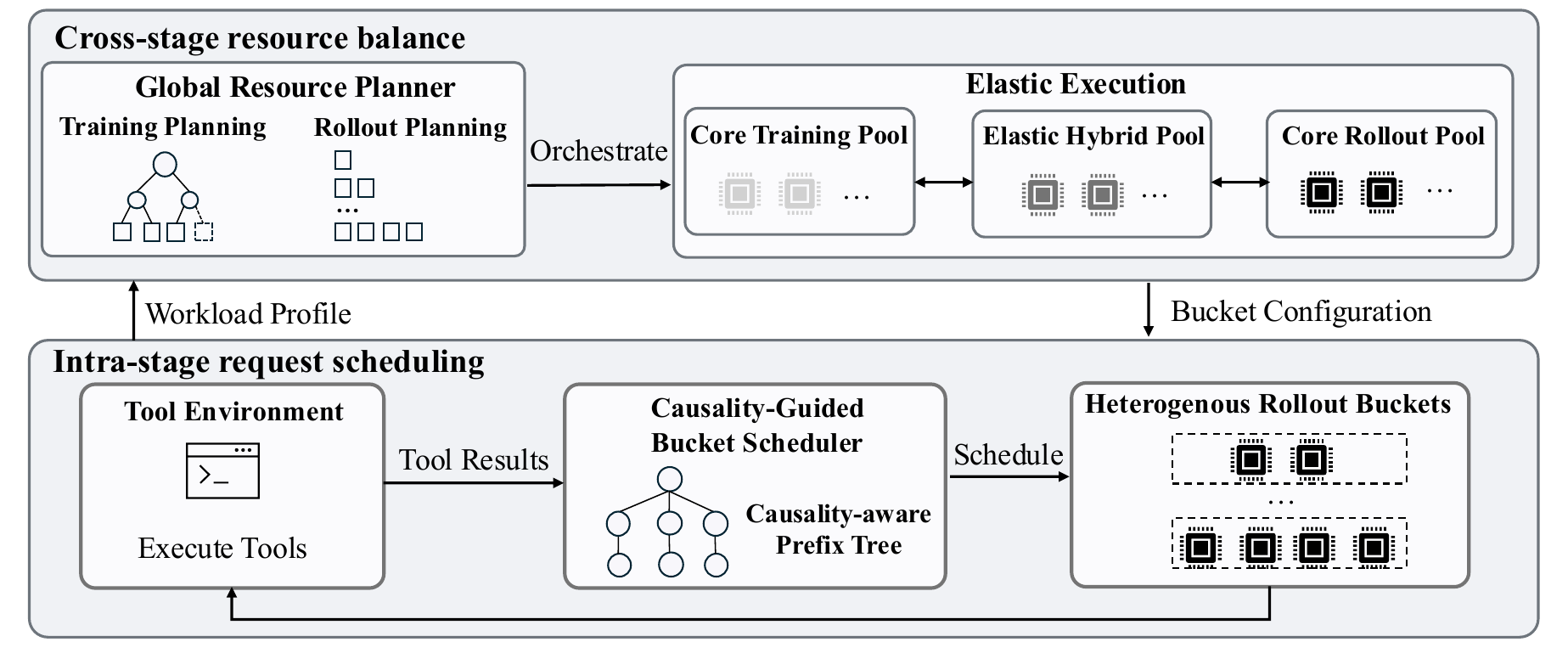}
        \caption{\sys overview}
        \label{sys_overview}
\end{figure}

\sys is an adaptive runtime for the disaggregated, asynchronous pipeline of agentic RL post-training.
 To address the long-tailed and cross-stage coordination challenges identified in previous sections, \sys organizes its execution around a two-level adaptive architecture (Figure~\ref{sys_overview}).

\noindent \textbf{Intra-Stage Scheduling (\S\ref{CGBS}).}
At the request level, \sys partitions the rollout cluster into heterogeneous TP buckets and routes each request to the bucket matching its predicted residual length. Routing is driven by the Causality-Guided Bucket Scheduler, which treats tool-execution outcomes as runtime signals for residual-length prediction, eliminating rollout stragglers without auxiliary predictors.

\noindent \textbf{Cross-Stage Coordination (\S\ref{sec:resource_planner}).}
Operating at the cluster level, cross-stage coordination maintains balance across the disaggregated pipeline.
A global planner periodically re-solves the cross-stage worker allocation and per-stage parallelism strategy under the current workload, guided by an analytical Cost Evaluator. 
Reallocations are enacted through two tiers: 
(i)~a lightweight tier where hybrid workers dynamically transition between rollout inference and training data-parallel replicas via decoupled communication domains and a non-blocking RDMA joining protocol, resolving transient bottlenecks without stalling ongoing training; and 
(ii)~a heavier tier that rebuilds the core pools' parallelism strategy during pipeline slack periods when the workload shift proves persistent.
\section{Global Resource Planner with Elastic Execution}
\label{sec:resource_planner}
In an asynchronous RL pipeline, the iteration time is bounded by the slower stage: $T_{\text{iter}} = \max(T_{\text{rollout}}, T_{\text{train}})$. 
The resource configurations of the training and rollout clusters are inherently coupled. 
Allocating more GPUs to training reduces $T_{\text{train}}$ but leaves fewer GPUs for rollout, increasing $T_{\text{rollout}}$ and restricting the range of viable heterogeneous TP configurations; 
symmetrically, over-provisioning the rollout cluster starves training of resources, raising $T_{\text{train}}$ and shifting the bottleneck to training. 
Consequently, local optimization leads to stage starvation. 
To maximize end-to-end throughput under a fixed GPU budget ($N_{GPU}$), the system must jointly optimize both clusters from a global perspective.

\sys addresses this coupled allocation problem with a hierarchical planner that decomposes the joint search space into training strategy selection and rollout partitioning to minimize the iteration makespan.
The training side employs a topology-aware decision tree with pruning to enumerate feasible parallel strategies, reducing the candidate set from exponential size to dozens.
The rollout side replaces exhaustive integer partitioning with a dynamic programming formulation.
This section details these two components in turn, and further describes how \sys realizes the planner's reallocations without global reconfiguration through the Elastic Hybrid Pool.
\begin{algorithm}[t]
    \caption{Global Resource Planner}
    \label{alg:hierarchical_planner}
    \footnotesize
    \begin{algorithmic}[1]
    \State \textbf{def} resourcePlanner($N_{GPU}$, $L$, $M_{limit}$, $\mathcal{T}$, $\text{CE}$):
    \State \quad $T^*_{\text{iter}} = \infty$
    \State \quad $\text{MemoRollout} = \emptyset$
    \State \quad \textbf{for} $n_{\text{train}} = 1$ \textbf{to} $N_{GPU}$ \textbf{do}
    \State \quad \quad $\mathcal{S}_{\text{cand}} = \text{DecisionTreeSearch}(n_{\text{train}}, M_{limit}, \text{CE})$
    \State \quad \quad \textbf{if} $\mathcal{S}_{\text{cand}} = \emptyset$ \textbf{then continue}
    \State \quad \quad $s_{\text{best}} = \operatorname{arg\,min}_{s \in \mathcal{S}_{\text{cand}}} \text{CE}.\text{TrainTime}(s)$
    \State \quad \quad $T_{\text{train}} = \text{CE}.\text{TrainTime}(s_{\text{best}})$
    \State \quad \quad $n_{\text{rollout}} = N_{GPU} - n_{\text{train}}$
    \State \quad \quad \textbf{if} $n_{\text{rollout}} \notin \text{MemoRollout}$ \textbf{then}
    \State \quad \quad \quad $\text{MemoRollout}[n_{\text{rollout}}] = \text{RolloutDP}(n_{\text{rollout}}, L, \mathcal{T}, \text{CE})$
    \State \quad \quad $T_{\text{rollout}} = \text{MemoRollout}[n_{\text{rollout}}]$
    \State \quad \quad $T_{\text{max}} = \max(T_{\text{train}}, T_{\text{rollout}})$
    \State \quad \quad \textbf{if} $T_{\text{max}} < T^*_{\text{iter}}$ \textbf{then}
    \State \quad \quad \quad $T^*_{\text{iter}} = T_{\text{max}}$, record $(n_{\text{train}}, s_{\text{best}})$
    \State \quad \textbf{return} optimal config
    
    \Statex
    \State \textbf{def} RolloutDP($n_{\text{rollout}}$, $L$, $\mathcal{T}$, $\text{CE}$):
    \State \quad Sort requests by length; let indices be $1 \dots L$
    \State \quad Init $dp[g][i] = +\infty$ for $0 \le g \le n_{\text{rollout}}, 0 \le i \le L$
    \State \quad $dp[0][0] = 0$
    \State \quad \textbf{for} $g = 1$ \textbf{to} $n_{\text{rollout}}$ \textbf{do}
    \State \quad \quad \textbf{for} $i = 0$ \textbf{to} $L$ \textbf{do}
    \State \quad \quad \quad \textbf{for} $tp \in \mathcal{T}$ \textbf{with} $tp \le g$ \textbf{do}
    \State \quad \quad \quad \quad \textbf{for} $x = 0$ \textbf{to} $i$ \textbf{do}
    \State \quad \quad \quad \quad \quad $t_{\text{inst}} = \text{CE}.\text{RolloutTime}(tp, i-x+1, i)$
    \State \quad \quad \quad \quad \quad $t_{\text{cand}} = \max(dp[g-tp][i-x], t_{\text{inst}})$
    \State \quad \quad \quad \quad \quad $dp[g][i] = \min(dp[g][i], t_{\text{cand}})$
    \State \quad \textbf{return} $dp[n_{\text{rollout}}][L]$
    \end{algorithmic}
    \end{algorithm}

\subsection{Training Side: Decision Tree-Based Parallel Strategy Enumeration}

The training strategy search space $\mathcal{S}$ comprises all valid $(TP, PP, DP)$ factorizations of $n_{\text{train}}$ for dense models. For MoE models, the space expands to $(TP, EP, PP, DP)$, where $EP$ denotes Expert Parallelism.
A naive enumeration of this space scales exponentially in the number of divisors of $n_{\text{train}}$.
\sys instead constructs a topology-aware decision tree that enumerates candidates level by level, pruning infeasible and inefficient branches at each depth.
The tree is extensible, using three levels for dense models and four levels for MoE models, with an additional $EP$ level inserted after $TP$.

\noindent \textbf{Decision tree structure.} The tree is rooted at $n_{\text{train}}$ and grows through parallel-strategy levels.
At the first level, $TP$ is restricted to $\{1, 2, 4, 8\}$, reflecting the single-node GPU count and the NVLink connectivity domain. Branches are pruned if the per-GPU memory footprint exceeds $M_{\text{limit}}$ or if the projected communication-to-computation ratio exceeds a configurable threshold $\alpha$.
For MoE models, the $EP$ level enumerates expert-parallel degrees that divide the total expert count and satisfy $TP \times EP \leq n_{\text{train}}$. Branches are pruned if their all-to-all communication volume exceeds a configurable threshold $\beta$, measured relative to the compute time of the corresponding MoE layers.
Placing $EP$ before $PP$ maximizes the likelihood of keeping all-to-all communication within a single node, which is critical for MoE performance.
At the next level, $PP$ candidates are integers satisfying $TP \times PP \leq n_{\text{train}}$ for dense models, or $TP \times EP \times PP \leq n_{\text{train}}$ for MoE models. Branches are pruned if their pipeline bubble ratio $\frac{PP-1}{PP+m-1}$, where $m$ denotes the number of micro-batches, exceeds a configurable threshold (e.g., 30\%).
At the final level, $DP = n_{\text{train}} / (TP \times PP)$ for dense models, or $DP = n_{\text{train}} / (TP \times EP \times PP)$ for MoE models. A path is valid only when this division yields an integer.
Each root-to-leaf path defines a complete training strategy $s_{\text{train}}$. \sys evaluates $T_{\text{train}}(s_{\text{train}})$ for every surviving strategy.

\subsection{Inference Side: Dynamic Programming for Optimal Heterogeneous Partitioning}

Given $n_{\text{rollout}}$ GPUs and a total workload of $L$ samples, the rollout subproblem partitions the GPUs into heterogeneous inference instances, each with a tensor-parallel degree $tp \in \mathcal{T}$ (where $\mathcal{T} = \{1, 2, 4, 8\}$), and assigns a subset of the workload to each instance. 
The objective is to minimize the makespan, i.e., the maximum completion time across all instances.
This problem exhibits optimal substructure and admits an efficient dynamic-programming solution, eliminating the need for brute-force enumeration over integer partitions.

\noindent \textbf{Cost interface.} The underlying cost evaluator provides a function $\text{Cost}(tp, a, b)$ that returns the minimum time for a single inference instance with tensor parallelism $tp$ to process the contiguous segment of requests indexed from $a$ to $b$ in the sorted list. 
This function internally accounts for maximum batch size limits, dynamic pipeline bubbles, and variable-length sequences, encapsulating all low-level execution complexity.

\noindent \textbf{DP formulation and solving procedure.}
Prior to the DP, all $L$ requests are sorted by their sequence length in non-decreasing order and indexed $1, 2, \dots, L$.
The DP constructs a heterogeneous partition by incrementally adding inference instances.
Each instance is characterized by a configuration $(tp, x)$, where $tp \in \mathcal{T}$ is its tensor-parallel degree and $x$ is the number of samples it serves.
Because different instances may take different $tp$ values, the resulting GPU partition forms a \emph{heterogeneous TP group}.
The DP enumerates all possible ways to append one such instance to an already-optimal sub-partition for the remaining GPUs and the remaining prefix of sorted samples, 
exploiting optimal substructure to solve the global makespan-minimization problem.

Concretely, let $dp[g][i]$ denote the minimum achievable makespan when using exactly $g$ GPUs to serve the first $i$ requests in the sorted list.
We initialize $dp[0][0] = 0$ and set $dp[g][i] = +\infty$ for all other states.
The state transition is:
{\footnotesize
\begin{align*}
    dp[g][i] &= \min_{\substack{tp \in \mathcal{T}, \ tp \le g \\ 0 \le x \le i}} \max\Big( dp[g-tp][i-x],\ \text{Cost}(tp, i-x+1, i) \Big)
\end{align*}
}
The transition considers the last added instance with configuration $(tp, x)$: it consumes $tp$ GPUs, processes the contiguous segment of requests indexed $i-x+1$ through $i$ in the sorted list, and contributes $\text{Cost}(tp, i-x+1, i)$ to the overall makespan.
The inner $\max$ captures the bottleneck instance time, while the outer $\min$ selects the best partition.
The optimal rollout time is $T^*_{\text{rollout}} = dp[n_{\text{rollout}}][L]$, and the optimal partition $P^*_{\text{rollout}}$ is recovered by backtracking from this state.

\noindent \textbf{Sample ordering by sequence length.}
Sequence length affects the optimal partition because large-$tp$ instances better accommodate long sequences, whereas small-$tp$ instances are more efficient for short sequences.
Sorting the requests by sequence length \emph{before} the DP traversal is essential for correctness: the DP state tracks how many requests from the sorted prefix have already been assigned, 
so every instance must receive a contiguous segment of the sorted list.
Consequently, the DP naturally dispatches shorter requests to smaller-$tp$ instances and longer requests to larger-$tp$ instances, making the length-to-TP matching both correct and effective.

\noindent \textbf{Length profiling.}
The sequence length of each prompt is drawn from historical rollout data.
Prior to training, an initial profiling run establishes a baseline length for every prompt.
Whenever a new rollout completes for a prompt, its stored length is updated with the latest observation, allowing the planner to adapt to distribution shifts over time.

\noindent \textbf{Complexity reduction.} The state space has size $O(n_{\text{rollout}} \cdot L)$. 
A naive transition enumeration over all $tp$ and $x$ would yield $O(n_{\text{rollout}} \cdot L^2 \cdot |\mathcal{T}|)$ complexity. 
This pseudo-polynomial complexity is still substantially smaller than exhaustive enumeration over all integer partitions of $n_{\text{rollout}}$, the partition function $p(n_{\text{rollout}})$ grows rapidly (e.g., $p(64) \approx 1.7 \times 10^6$). 
In practice, the rollout subproblem remains tractable because $|\mathcal{T}|$ is small and each rollout DP result is memoized for reuse by outer-loop states with the same rollout budget.

\subsection{Hierarchical Search with Rollout Memoization}

Algorithm~\ref{alg:hierarchical_planner} integrates the two subproblem solvers into a global nested search. 
The outer loop iterates over feasible values of $n_{\text{train}}$ (e.g., $1$ to $N_{GPU}$) and invokes the decision tree to obtain the set of valid training strategies $\mathcal{S}_{\text{cand}}$ for each budget, yielding the optimal training time $\min_{s \in \mathcal{S}_{\text{cand}}} T_{\text{train}}(s)$. 
The rollout DP is invoked once per distinct $n_{\text{rollout}} = N_{GPU} - n_{\text{train}}$ and its result is memoized in $\text{MemoRollout}$. 
Because different outer-loop states that share the same rollout budget reuse the memoized result, the planner avoids redundant DP executions and remains efficient even at large cluster scales.
In Algorithm~\ref{alg:hierarchical_planner}, $\text{CE}$ denotes the cost evaluator, which provides the estimated training and rollout times used by the planner; its design is detailed in \S\ref{section:CE}.

The planner is invoked periodically (e.g., every $K$ training iterations) rather than once at initialization. 
The workload $L$ is continuously updated as prompts are revisited during rollout; at each invocation, the planner reads the latest statistics and computes the new optimal configuration. 
If the projected throughput gain exceeds the reconfiguration cost, \sys triggers a resource reallocation. 

\begin{figure}[t]
    \centering
        \includegraphics[width=\columnwidth]{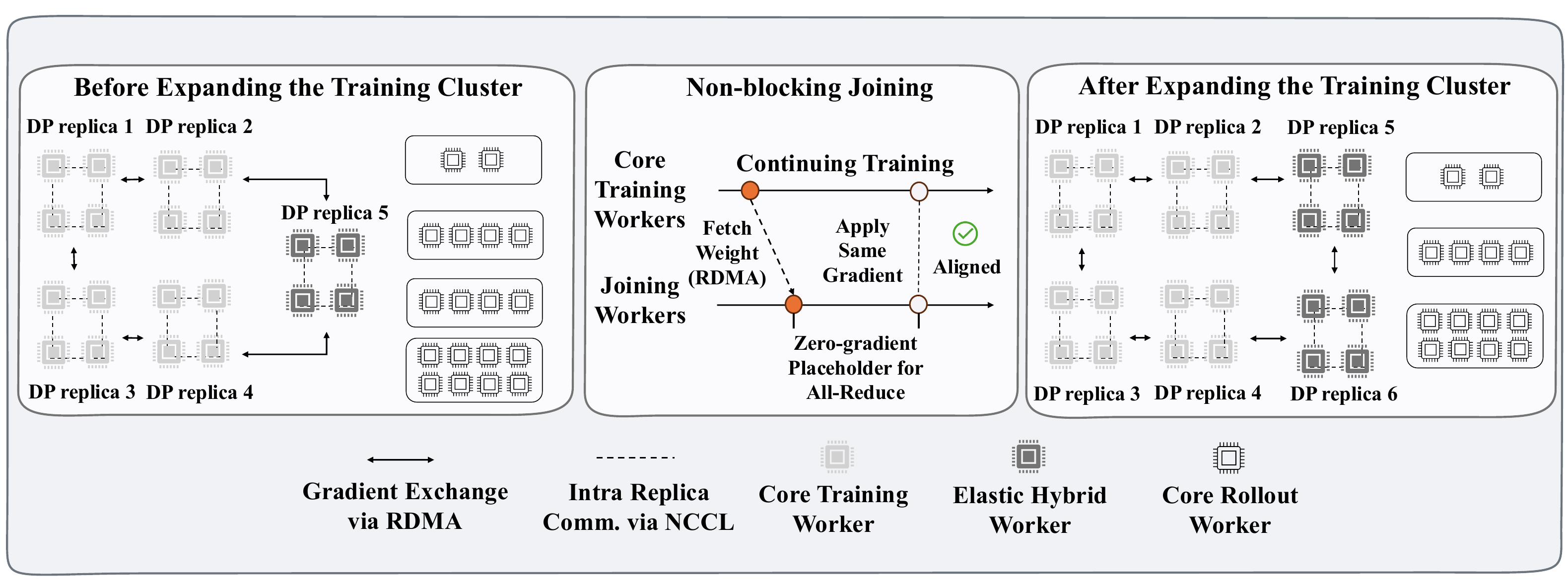}
        \caption{Elastic execution example.}
        \label{design1}
\end{figure}

Realizing these reallocations, however, requires bridging a gap between planning and execution: in conventional frameworks~\cite{megatron,fsdp,fsdp2}, adding or removing training workers forces a global communicator rebuild and state redistribution, making frequent reallocation impractical.
Figure~\ref{design1} illustrates an example of how \sys expands the training cluster by adding a hybrid worker without disrupting the core training pool.

\subsection{Elastic Hybrid Pool}
\label{section:elastic_hybrid_pool}
\sys's insight is that resource elasticity need not be a global reconfiguration event. 
If the core training topology is kept strictly immutable and workers enter or exit only as complete, unsharded data-parallel replicas, 
they can participate as \emph{external} members rather than \emph{internal} collective ranks.

\noindent \textbf{Three-pool organization.}
To materialize these principles, \sys partitions the cluster into a Core Training Pool, a Core Rollout Pool, and an Elastic Hybrid Pool.
The two core pools provide stable capacity for the dominant workload modes.
The Core Training Pool performs synchronized weight updates and maintains static data-, tensor-, and pipeline-parallel groups.
The Core Rollout Pool handles trajectory generation with the heterogeneous TP configurations determined by the planner.
The Elastic Hybrid Pool provides fast-response capacity for resource reallocation.
When rollout becomes the bottleneck, Hybrid workers switch to rollout mode to expand generation throughput.
Conversely, when training becomes the bottleneck, they rejoin training as additional DP replicas.
Because each Hybrid worker holds a complete, unsharded copy of model weights and optimizer states, its addition or removal does not reshape the tensor- or pipeline-parallel topology of the core training pool.
Importantly, the Elastic Hybrid Pool need not be fully reserved in advance:
because rollout instances are stateless, \sys can temporarily borrow workers from the Core Rollout Pool, convert them into Hybrid workers, and later return them with low transition cost.
This mechanism covers transient training-side shortages without perturbing the core training topology.
Rollout-side shortages are handled even more cheaply by switching existing Hybrid workers into rollout mode.
If RL optimization induces a persistent demand shift, \sys can rebuild the core partition itself;
such rebuilds are infrequent, because short-term imbalances are absorbed by the Elastic Hybrid Pool and large repartitioning can be deferred to slack periods---intervals in which training is already stalled waiting for fresh rollout data, which naturally serve as reconfiguration windows.

\noindent \textbf{Decoupled communication domains.}
Building on non-blocking fault-tolerant training~\cite{FT-HSDP},
 \sys decouples intra-replica and inter-replica communication into independent planes.
 We define the intra-replica domain as the NCCL collective within a single data-parallel replica; this domain remains strictly stable and is never rebuilt.
The inter-replica domain, conversely, handles traffic across data-parallel replica (e.g., gradient exchange between the core training pool and elastic hybrid workers).
All elasticity-related membership changes are confined to the inter-replica domain, where an independent control plane manages topology rewiring and RDMA connection establishment without disturbing the data plane.
Implementation details are provided in Appendix~\ref{appendix:decoupled_comm}.

\noindent \textbf{Non-blocking joining.}
Transitioning a Hybrid worker from rollout back to training requires reloading model and optimizer states.
If the active training cluster were to pause and wait for this restoration, it would suffer from severe synchronization overhead.
\sys avoids this pause through a non-blocking joining protocol inspired by non-blocking fault-tolerant training~\cite{FT-HSDP}.
At the end of each training step, active training workers asynchronously capture a snapshot of their current weights and optimizer states.
When a Hybrid worker transitions back to training, it fetches this snapshot and joins as a DP replica.
During the restoration window, the active training cluster advances to the next step without waiting.

Crucially, the joining worker does not directly join the core training pool's All-Reduce collective.
Instead, it routes its gradient to the corresponding core training rank through the asynchronous side channel via RDMA.
During recovery, this gradient is a zero placeholder.
Because the core rank accumulates the external gradient into its local backward pass, a zero placeholder leaves the core rank's local gradient unchanged.
Consequently, the intra-core All-Reduce produces exactly the same averaged gradient as training without the joining worker, preserving mathematical equivalence (Appendix~\ref{appendix:rejoin}).
Once the snapshot is loaded, the joining worker's parameters and optimizer state are overwritten by the snapshot, attaining identical state to the rest of the cluster without requiring iteration replay or global synchronization.

\subsection{Underlying Cost Evaluator}
\label{section:CE}
The planner relies on an accurate cost evaluator (CE) to provide $\text{CE}.\text{TrainTime}(s)$ and $\text{CE}.\text{RolloutTime}(tp, a, b)$. 
However, existing simulators need to be adapted for RL post-training because the highly variable length of generated sequences causes severe workload imbalance. 
Inference simulators~\cite{vidur} emphasize general request serving and model operator execution via black-box estimation, which lacks the generalization capability needed for the extreme length variations in RL generation. 
Training simulators~\cite{sailor,galvatron} rely on a steady-state assumption, 
calculating a static pipeline beat based on uniform micro-batch lengths. 
This simplification misses the dynamic pipeline bubbles where downstream stages are forced to idle while waiting for the completion of a micro-batch containing long-tail sequences.
To accurately reflect the physical execution complexity in RL post-training scenarios, \sys's cost evaluator inherits mechanisms from existing simulators and modifies them to support dynamic length distributions.
Please note that the CE is a modular design; other simulators~\cite{moye,simai,frontier} can also be adapted with modifications.

For the rollout phase, \sys inherits profiling-guided, operator-triaged runtime prediction from Vidur~\cite{vidur}. 
However, it replaces Vidur's random forest estimators with polynomial fitting---$O(L)$ for Linear operators and $O(L^2)$ for Attention operators. 
This replacement is necessary because polynomial functions provide stronger physical interpretability and robust generalization for the extreme sequence length variations in RL generation. 
Specifically, it formulates the execution time of each layer and stage as a direct function of the sequence length $L$, 
such as $T_{\text{Attn}}(L) = \gamma \cdot L^2 + \eta \cdot L + \theta$, rather than relying on black-box predictions like random forests.
The $\text{CE}.\text{RolloutTime}(tp, a, b)$ interface used by the DP internally invokes these per-operator models, aggregates them across layers and pipeline stages, and accounts for batching effects and dynamic load imbalance across the assigned segment of samples.

For the training phase, \sys inherits the execution graph simulation (e.g., 1F1B Pipeline Parallelism) and iteration time prediction framework from Sailor~\cite{sailor}. 
The primary modification is abandoning Sailor's steady-state assumption in favor of micro-bubble capture.
This shift is required because the heterogeneous length distribution creates misaligned computational loads across micro-batches, 
meaning a uniform steady-state beat simply does not exist. 
Instead of multiplying a bottleneck stage duration by the number of micro-batches, \sys explicitly calculates the start and finish times for each micro-batch at each pipeline stage, taking into account data arrival and hardware availability. 
This non-uniform simulation accurately captures the bottleneck drift and accumulated micro-bubbles caused by length mismatch.

\section{Causality-Guided Bucket Scheduler for Heterogeneous Rollout Clusters}
\label{CGBS}
\begin{figure}[t]
    \centering
        \includegraphics[width=\columnwidth]{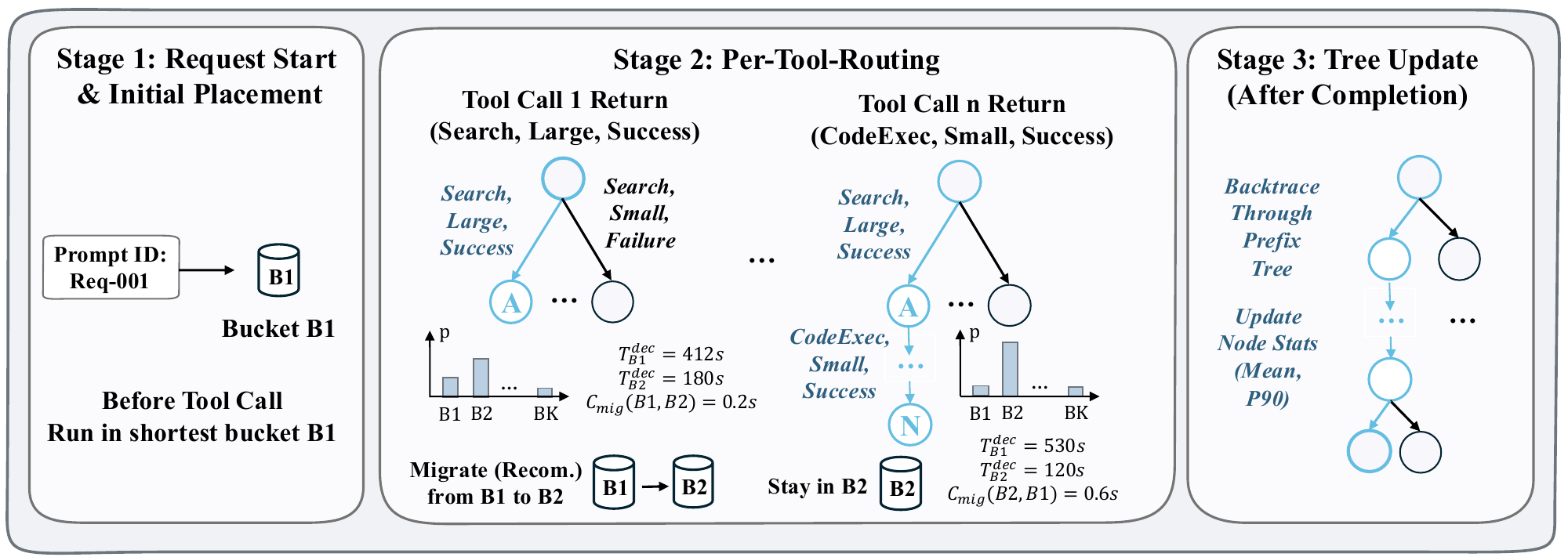}
        \caption{A scheduling example of Causality-Guided Bucket Scheduler.}
        \label{design2}
    \end{figure}

We first present \sys's intra-stage scheduling, which addresses length heterogeneity and eliminates stragglers through Causality-Guided Bucket Scheduler across heterogeneous execution buckets.
As motivated in Section~\ref{section:opportunity}, \sys instantiates the rollout cluster as a set of heterogeneous buckets $\mathcal{B} = \{B_1, B_2, \dots, B_m\}$ ordered by increasing compute capacity, 
where each bucket contains inference instances with a distinct TP configuration. 
The global planner described in \S\ref{sec:resource_planner} provisions these buckets, and the bucket scheduler serves as the runtime scheduling policy that maps individual requests to them conditioned on the causal signals revealed by tool execution.
Figure~\ref{design2} illustrates the end-to-end Causality-Guided Bucket Scheduler workflow on a heterogeneous rollout cluster.

\noindent \textbf{Causality-aware prefix tree.}
To turn this causal intuition into a routing decision, \sys builds a lightweight prefix-tree (trie) structure from historical trajectories. 
Each node in the tree represents a unique prefix of the tool-call history, keyed by the ordered sequence of return states from all prior tool interactions. 
An edge corresponds to invoking a specific tool type and receiving its abstracted return state, which encodes the payload characteristics (e.g., size class) and the execution outcome (e.g., success or failure). 
Formally, a node at depth $k$ is uniquely identified by the prompt identifier and the sequence of $k$ observed return states, denoted $\kappa = (\texttt{prompt\_id}, \langle s_1, s_2, \dots, s_k \rangle)$, where each $s_i$ denotes the composite state of the $i$-th tool call, 
comprising the tool type invoked as well as its return attributes such as payload size ($\textsc{SizeSmall}$, $\textsc{SizeLarge}$) and execution status ($\textsc{ToolSuccess}$, $\textsc{ToolFailure}$), derived from the environment response.
Every node stores a histogram of the remaining sequence length distribution from that point onward, with bins aligned to the length thresholds of the buckets, for runtime routing. 
\sys builds the tree offline from trajectory logs (Appendix~\ref{appendix:profile}): for each trace, it walks down the tree and records the remaining length at every visited node, enabling $O(1)$ runtime lookups of conditional residual length distributions.

\noindent \textbf{Three-stage scheduling pipeline.}
Causality-Guided Bucket Scheduler uses the prefix tree to route requests through three phases.

\textit{Phase 1: Initial placement.}
Before the first tool call, all trajectories are placed in the shortest bucket. 
Because the initial prompt and early reasoning contents typically occupy limited context, 
assigning every request to a high-TP instance would reduce cluster-wide utilization.

\textit{Phase 2: Per-tool-return routing.}
When the model emits a tool-call token, \sys pauses decoding, offloads the request from the GPU, and waits for the external environment to execute the tool asynchronously.
After the tool returns, the scheduler extracts the return-state label and walks down the prefix tree from the current node.
At the reached node, it retrieves the empirical histogram $F = \{p_1, p_2, \dots, p_K\}$ of the remaining trajectory length, where $p_k$ is the probability mass in bin $k$. The bins are aligned with the buckets's thresholds;
Causality-Guided Bucket Scheduler then casts routing as a cost-benefit decision: for each candidate bucket $B \in \mathcal{B}$, it estimates the expected remaining decode time as $T^{\text{dec}}_B = \sum_{k=1}^{K} p_k \cdot \bar{T}_{\text{dec}}(B,k)$, where $\bar{T}_{\text{dec}}(B,k)$ is the decode cost for bucket $B$ on bin $k$ (using the planner's cost model, described in \S\ref{section:CE}), and selects
\begin{equation*}
B^* = \arg\min_{B \in \mathcal{B}} \Big( \sum_{k=1}^{K} p_k \cdot \bar{T}_{\text{dec}}(B,k) + C_{\text{mig}}(B_{\text{cur}}, B) \Big),
\end{equation*}
where $C_{\text{mig}}(B_{\text{cur}}, B)$ is the cost of moving the request's KV state to $B$, and $C_{\text{mig}}(B_{\text{cur}}, B_{\text{cur}}) = 0$ . It is profiled per network path as the cheaper of transfer and recomputation (detailed below).
If the walk reaches an unseen node, the scheduler backs off to its parent's distribution.

\textit{Phase 3: Tree update.} 
\label{phase3}
Offline, \sys incrementally updates the tree by inserting newly collected traces along their return-state sequences (Appendix~\ref{appendix:subsec:update}), recording residual lengths at visited nodes to continuously align empirical distributions with policy drift.

This design yields three advantages: routing decisions require only cheap $O(1)$ lookups without online model inference, policy drift is tracked via lightweight offline tree updates without parametric model retraining, and requests migrate directly to target buckets in a single step upon tool returns rather than through incremental promotion.

\noindent \textbf{Migration cost and system behavior.}
\sys migrates a request across buckets in three steps, each of which incurs at most negligible overhead.

\textit{Step 1: KV-cache offload during tool execution.}
When the model emits a tool-call token, the source GPUs offload the request's KV cache to CPU pinned memory through PCIe.
This offload is not migration-specific: it is standard practice in production inference systems to offload a paused request's KV cache to CPU memory so that concurrent decoding requests can proceed efficiently~\cite{mori,tokencake,Continuum}.
\sys performs the same offload whether the request will later migrate or resume in place, so the PCIe transfer time can be masked by the asynchronous tool I/O and never appears on the scheduling critical path.

\textit{Step 2: CPU-side resharding.}
While the KV cache resides in CPU memory, \sys reshards it along the attention-head dimension to match the target bucket's TP configuration.
If the target TP is larger ($\text{TP}_{\text{target}} > \text{TP}_{\text{source}}$), each source partition is split into $\frac{\text{TP}_{\text{target}}}{\text{TP}_{\text{source}}}$ shards;
if the target TP is smaller, every $\frac{\text{TP}_{\text{source}}}{\text{TP}_{\text{target}}}$ contiguous partitions are concatenated into one.
Both operations are pure CPU memory rearrangements with no data movement across buses, so their latency is negligible.

\textit{Step 3: GPU reload at tool return.}
When the tool result returns and the target bucket is determined, each target GPU loads its own shard from CPU memory and resumes decoding.
If the source and target buckets reside on the same node, the reload latency is essentially identical whether the request moves to another GPU or returns to the same one.

For cross-node migration, the KV cache must traverse the inter-node network before decoding can resume.
\sys therefore compares the end-to-end cost of transferring the KV cache against the cost of recomputing it from scratch on the target bucket. 
This comparison accounts for the full migration pipeline: PCIe staging, CPU-side resharding, network transfer, and target-GPU reload. 
\sys builds the decision on empirically measured latency profiles. 
Offline micro-benchmarks characterize the migration latency (including all staging overheads) 
and the recomputation prefill latency for each target TP configuration across 
the relevant range of sequence lengths. 
At runtime, \sys queries this profile to select the cheaper path for each request. 
As shown in Section~\ref{sec:overhead}, the measured end-to-end migration latency stays below 733~ms even at 40K-token prefixes,
and the crossover point where recomputation becomes cheaper falls around 4K tokens for typical configurations.

\section{System Implementation}
\sys is implemented in approximately 13,000 lines of Python and C++/CUDA code.
The core asynchronous RL training pipeline is built on top of verl~\cite{verl} (v0.8.0), using vLLM~\cite{vllm} (v0.9.2 on CUDA; v0.22.1 on Ascend NPU) for generation and Megatron-Core~\cite{megatron} (v0.14.0, with \texttt{megatron-bridge} v0.2.0rc6) for policy training under PyTorch 2.7.0 (torch-npu 2.7.1). 
We further describe the implementation of some of the unique aspects of \sys.

\noindent \textbf{Decoupled communication domains.}
To enable elastic execution, \sys separates control plane and data plane in cross-replica gradient exchange. 
The CPU manages all complex control logic: adding/removing replicas, 
RDMA connection re-establishment, timeout handling, and congestion control via fixed-size chunk pipelining (16 MB chunks). 
The GPU only executes data copy and reduction kernels using a ring algorithm (ReduceScatter + AllGather). 
This hybrid design allows us to reconfigure communication groups without restarting core training pool. 
The CPU thread coordinates with a consensus service to detect failures and determine the active replica set; 
then it updates RDMA send/receive buffers accordingly. 
Meanwhile, the GPU stream busy-polls a host-pinned flag—when data arrives, 
it performs in-GPU reduction. 
The kernel is optimized to use only 2 SMs (vs. NCCL's 4) by exploiting instruction-level parallelism and register reduction, achieving comparable bandwidth while avoiding contention with computation (Appendix~\ref{appendix:decoupled_comm}).

\noindent \textbf{Non-blocking joining.}
Elastic workers fetch the latest checkpoint directly from the core training pool via a load-balanced P2P RDMA protocol over CPU host memory,
which runs over host memory and does not interfere with GPU high-speed network.
To join without stalling, \sys leverages a zero-gradient trick:
the recovering replica sends a zero gradient through its side channel to the corresponding core rank instead of its computed one.
Because the core rank's local gradient is unchanged by this zero placeholder, the subsequent intra-core All-Reduce remains mathematically equivalent to training without the recovering replica.
Once the snapshot is loaded, all replicas resume from identical parameter and optimizer states (Appendix~\ref{appendix:rejoin}).

\section{Evaluation}
\label{evaluation}
\begin{figure*}[h!]
    \centering
        \includegraphics[width=\textwidth]{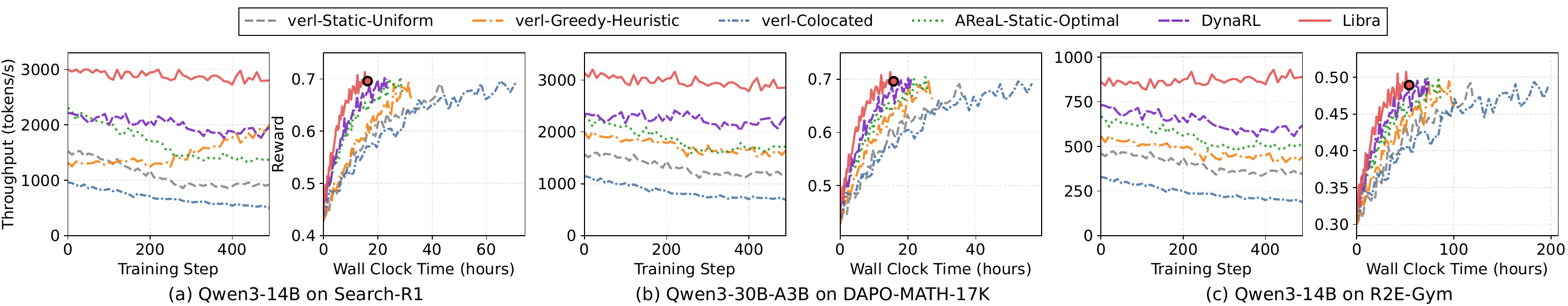}
        \includegraphics[width=\textwidth, trim=0 0 0 36, clip]{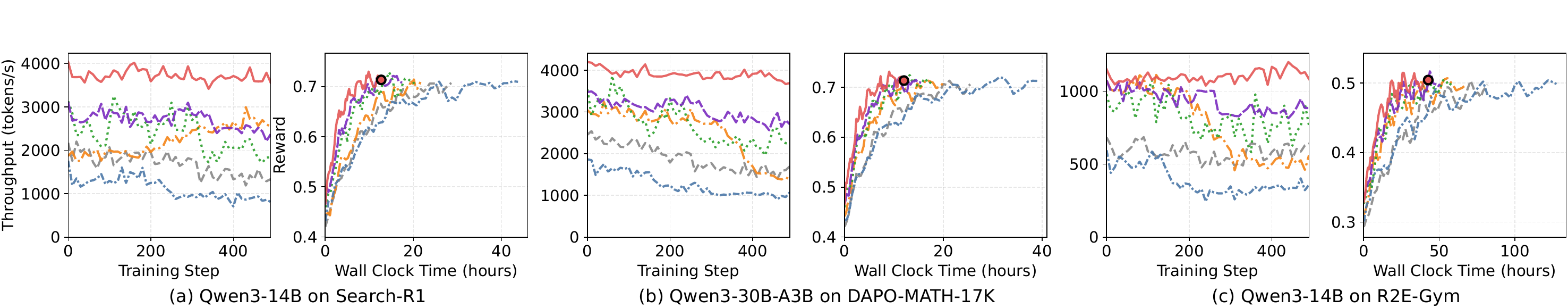}
        \caption{End-to-end training performance on three benchmarks. Each column shows throughput (left) and reward convergence (right) for one workload. Top row: GPU cluster; bottom row: NPU cluster.}
        \label{fig:main_exp}
\end{figure*}
\textbf{Setup.}
We deploy \sys on two clusters. 
The first is a GPU cluster of 48 NVIDIA A800-SXM4-80GB GPUs (6 nodes, 8 GPUs per node), connected via NVLink (NVSwitch, 600 GB/s bidirectional) and a 200 Gb/s RoCE (RDMA over Converged Ethernet) fabric with GPUDirect RDMA.
The second is an NPU cluster of 160 Ascend 910B3 NPUs (20 nodes, 8 NPUs per node, 64 GB HBM each). 
Intra-node NPUs are fully interconnected through HCCS: each NPU connects to the other seven via seven 224-Gb/s HCCS links, yielding 196 GB/s of aggregate unidirectional intra-node bandwidth per NPU.
Inter-node communication uses a 200 Gb/s NPU RoCE fabric.
All end-to-end experiments use the GRPO algorithm, a maximum model length of 40960 tokens, and 16 samples per prompt.
Following prior work\cite{Laminar,OrchestrRL,streamrl,areal}, experiments adopt the one-step asynchronous setting.
While our evaluation focuses on GRPO, \sys is algorithm-agnostic and can be extended to other RL algorithms such as PPO\cite{ppo}.

\noindent \textbf{Models.}
We conduct experiments on two widely adopted open-weight models: Qwen3-14B and Qwen3-30B-A3B (30B total parameters with 3B activated per token, a Mixture-of-Experts architecture)~\cite{qwen3}.

\noindent \textbf{Baselines.}
We compare \sys against five representative baselines:
\begin{itemize}[nosep,leftmargin=*]
\item \textit{verl-Colocated} replicates the widely used execution mode adopted by existing RL frameworks~\cite{verl,roll,slime,openrlhf,deepspeed}.
In this mode, training and rollout share the same set of GPUs and execute alternately through the hybrid engine without explicit partitioning. All three \textit{verl} baselines are deployed on the same verl codebase (v0.8.0) with identical Megatron-Core (v0.14.0) and vLLM (v0.9.2) backends as \sys.
\item \textit{verl-Static-Uniform} evenly splits the cluster into two halves: one for training and one for rollout. This configuration is adopted in a subset of experiments by recent disaggregated RL systems~\cite{Laminar,OrchestrRL}.
\item \textit{verl-Greedy-Heuristic} minimizes training GPUs subject to memory and batch constraints, assigning remaining GPUs to high-TP rollout instances.
\item \textit{AReaL-Static-Optimal} represents AReaL~\cite{areal,real} using its official open-source codebase (v2.0.0), configured with the best static partition and parallelism strategy identified by \sys's cost evaluator~(\S\ref{section:CE}).
\item \textit{DynaRL}~\cite{DynaRL} dynamically shifts GPU counts between stages at runtime. We evaluate DynaRL using its official implementation in RLinf~\cite{rlinf} (v0.3). 
\end{itemize}

\noindent \textbf{Workloads.}
We evaluate \sys on three representative agentic RL benchmarks:
(1)~\textit{Search-R1}~\cite{searchr1}, an information-retrieval benchmark with multi-turn search interactions;
(2)~\textit{R2E-Gym}~\cite{r2egym}, a software-engineering environment with 8.1K problems where agents invoke Bash/Python tools to edit code and verify patches; and
(3)~\textit{DAPO-Math-17K}~\cite{DAPO}, a mathematical reasoning dataset of 17K competition-level problems.
\subsection{End-to-End Experiments}

Figure~\ref{fig:main_exp} compares \sys against five baselines across all workloads on the GPU (48$\times$ A800) and NPU (160$\times$ Ascend 910B3) clusters.
Each column presents one workload, plotting throughput (left) and task reward versus wall-clock time (right).

\noindent \textbf{Throughput.}
\sys consistently delivers the highest throughput across all benchmarks on both platforms.
On the GPU cluster, \sys achieves 870--2,940~token/s, outperforming the strongest dynamic baseline DynaRL by 31\%--44\% (e.g., 2,900 vs.~2,000~token/s on Search-R1), AReaL-Static-Optimal by 57\%--73\%, and verl-Colocated by 246\%--321\% ($3.5\times$--$4.2\times$ throughput speedup).
On the NPU cluster, \sys sustains 1,100--3,900~token/s, exceeding DynaRL by 18\%--37\% and static/colocated baselines by 33\%--233\%.

\noindent \textbf{Reward convergence.}
All methods reach comparable target rewards; the primary distinction is the wall-clock time required to get there (marked by the red dot for \sys in Figure~\ref{fig:main_exp}).
\sys finishes first on every benchmark on both clusters.
Compared to DynaRL, \sys achieves $1.31\times$--$1.45\times$ speedups across the three benchmarks on GPUs (e.g., reaching target reward in 16.2~h vs.~23.5~h on Search-R1, and 54.0~h vs.~73.8~h on R2E-Gym) and $1.18\times$--$1.37\times$ speedups on NPUs (12.7~h vs.~17.4~h on Search-R1, and 42.7~h vs.~50.5~h on R2E-Gym).
Against verl-Static-Uniform, \sys accelerates end-to-end training by up to $2.73\times$ on GPUs and $2.26\times$ on NPUs.

\subsection{Effectiveness of Causality-Aware Routing}

To isolate the impact of routing quality, we compare \sys's bucket scheduler against four alternatives on the Search-R1 workload with the same heterogeneous bucket configuration.
\textit{Oracle} routes each request using the ground-truth remaining sequence length known at runtime; it represents an unattainable upper bound since the true length is unavailable in advance.
\textit{Prediction-Based} employs the embedding-based length predictor from ~\cite{embedding}, which estimates sequence lengths from hidden-state embeddings.
Under RL training, however, the LLM is continuously updated, causing the predictor's embeddings to drift and accuracy to degrade without periodic retraining.
\textit{MLFQ} adopts a reactive policy that migrates a request only after its length exceeds the current bucket's upper bound, inevitably incurring late and frequent migrations.
\textit{Load Balancing} distributes requests uniformly across buckets, ignoring sequence length.

Table~\ref{tab:routing_comparison} reports per-decision routing accuracy, migration ratio, and system throughput, while Figure~\ref{fig:cmlfq_ablation} traces throughput evolution across training steps.
Per-decision accuracy measures the probability that a scheduling algorithm makes a choice consistent with the ground-truth optimal bucket at each scheduling decision point.
Causality-Guided Bucket Scheduler achieves 91.1\% per-decision accuracy, closing most of the gap to the Oracle (100\%) and substantially outperforming Prediction-Based (65.2\%), MLFQ (44.8\%), and Load Balancing (31.2\%). 
As for system throughput: \sys delivers 2,700 tokens/s, outperforming Prediction-Based, MLFQ, and Load Balancing by 18\%, 41\%, and 80\%, respectively. Figure~\ref{fig:cmlfq_ablation} corroborates this trend: \sys sustains 2,500--2,800 tokens/s across training steps, closely tracking the Oracle curve and consistently exceeding the other policies by 400--1,000 tokens/s.

\begin{table}[t]
    \centering
    \begin{minipage}[t]{0.5\columnwidth}
        \centering
        \vspace{0pt}
        \scriptsize
        \setlength{\tabcolsep}{1.4pt}
        \resizebox{\textwidth}{!}{
            \begin{tabular}{@{}l r r l@{}}
            \toprule
            \makecell[c]{Method} & \makecell[c]{Per Decision\\Acc\%} & \makecell[c]{Migrate\\Ratio\%\textsuperscript{*}} & \makecell[c]{Throughput\\(token/s)}\\
            \midrule
            \makecell[c]{Load Balancing} & \makecell[c]{31.2} & \makecell[c]{0} & \makecell[c]{1502}  \\
            \makecell[c]{MLFQ} & \makecell[c]{44.8} & \makecell[c]{108} & \makecell[c]{1917} \\
            \makecell[c]{Prediction-Based} & \makecell[c]{65.2} & \makecell[c]{48.9} & \makecell[c]{2289} \\
            \makecell[c]{Bucket Scheduler} & \makecell[c]{91.1} & \makecell[c]{8.2} & \makecell[c]{2700}\\
            \makecell[c]{Oracle} & \makecell[c]{100} & \makecell[c]{0} & \makecell[c]{3102}\\
            \bottomrule
            \end{tabular}
        }
        {\scriptsize *Migration Ratio $= \dfrac{\text{total migrated tokens}}{\text{total tokens}}$.}
        \caption{Comparison of routing policies.}
        \label{tab:routing_comparison}
    \end{minipage}
    \hfill
    \begin{minipage}[t]{0.46\columnwidth}
        \centering
        \vspace{0pt}
        \includegraphics[width=\textwidth]{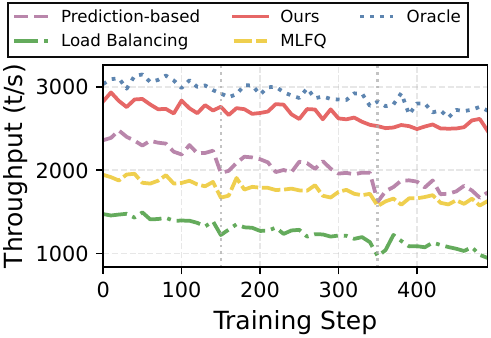}
        \vspace{-7mm}
        \captionof{figure}{Ablation study of the scheduler.}
        \label{fig:cmlfq_ablation}
    \end{minipage}
\end{table}

\subsection{Ablation Study}

\begin{figure}[t]
    \centering
        \includegraphics[width=\columnwidth]{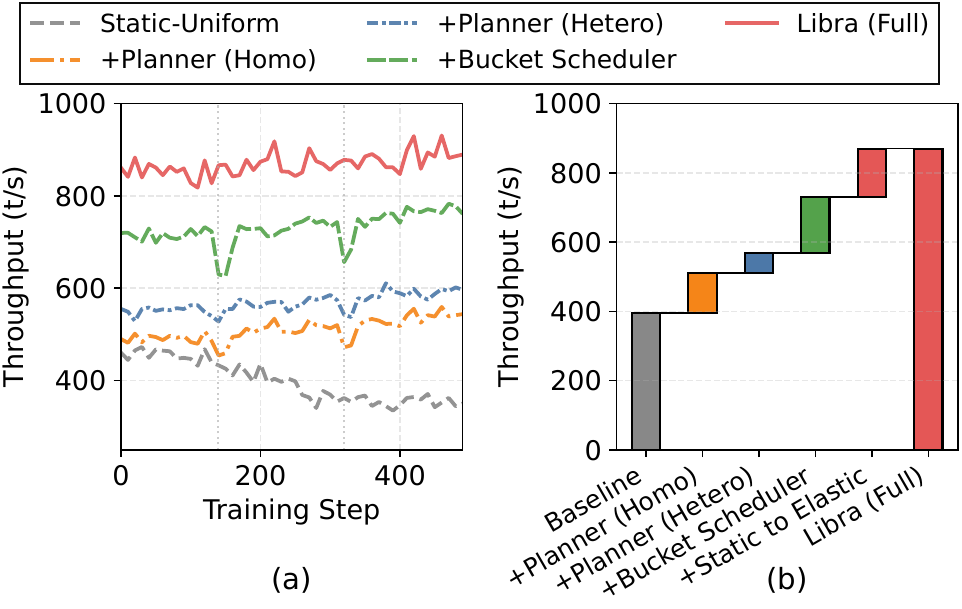}
        \caption{Ablation study on R2E-Gym (Qwen3-14B).
        (a)~Throughput over training steps for progressively enabled \sys components.
        (b)~Waterfall breakdown of average throughput gains from each component.}
        \label{fig:r2e_gym_ablation}
\end{figure}

Figure~\ref{fig:r2e_gym_ablation} quantifies the incremental throughput gains from progressively enabling \sys's four core components on R2E-Gym (Qwen3-14B).
Panel~(a) shows throughput evolution across training steps, and panel~(b) presents the waterfall breakdown of average throughput contributions.
Starting from the verl-Static-Uniform baseline (396~token/s), enabling the Planner with homogeneous TP raises throughput to 511~token/s ($\sim29.0\%$) by better balancing rollout and training.
Heterogeneous TP support adds a further 57~token/s ($\sim11.2\%$).
The Causality-Guided Bucket Scheduler contributes the largest gain, 162~token/s ($\sim28.5\%$), by routing requests to size-appropriate TP buckets.
Finally, Elasticity yields a 140~token/s gain ($\sim19.2\%$), dynamically shifting idle workers to the bottleneck stage.
In total, \sys reaches 870~token/s ($\sim120\%$ above the baseline), matching the steady-state throughput reported in the end-to-end evaluation.
Panel~(a) also shows that \sys recovers fastest from workload drift, confirming that elasticity is essential for sustained performance under shifting conditions.

\subsection{Overhead Analysis}
\label{sec:overhead}
Table~\ref{tab:worker_transition} breaks down worker-state transition costs for Qwen3-14B on R2E-Gym.
Switching from training to rollout incurs 15~ms teardown and 10.8~s vLLM activation; because weights reside on GPU and CUDA graphs are reused, this is an amortized one-time warm-up.
The reverse transition (rollout$\rightarrow$training) overlaps a 3.6~s snapshot with training, reloading state via RDMA in 4.4~s with $\lesssim$40~ms gradient and state alignment overhead.
Crucially, transitions are triggered only during reconfiguration rather than every step.
Across a 500-step Search-R1 run (Appendix~\ref{appendix:rejoin}), hybrid transitions occurred only $\sim$26 times (once every 15--20 steps).
Compared to the 116.6~s average step time, the 4.4~s reload is hidden behind active training passes, keeping cumulative transition overhead below 0.3\% of the 16.2-hour wall-clock time.
Furthermore, a heavier core partition rebuild was triggered only once at step 252, \sys deferred this rebuild to a natural pipeline slack period (while training was awaiting fresh rollout batches), the reconfiguration time was absorbed within the pipeline bubble.  

\begin{figure}[t]
    \centering
        \includegraphics[width=\columnwidth]{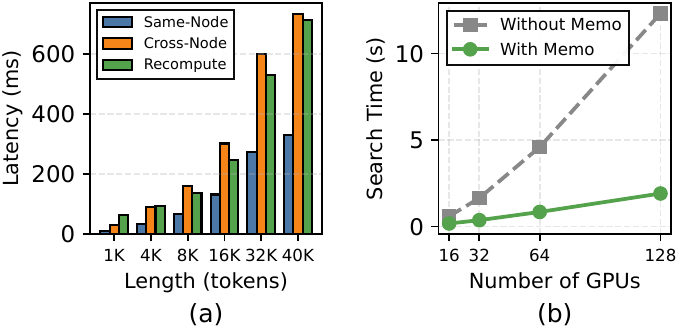}
        \caption{Overhead analysis.
        (a)~KV cache migration latency across sequence lengths.
        (b)~Planner search time vs.~GPU count.}
        \label{fig:overhead_analysis}
    \end{figure}

Figure~\ref{fig:overhead_analysis} quantifies remaining runtime overheads.
Panel~(a) shows KV-cache migration latency versus sequence length.
Same-node transfer stays below 330~ms even at 40~K tokens, while cross-node transfer reaches 733~ms.
Recomputation becomes cheaper than cross-node migration at longer sequences.
Panel~(b) measures planner search time as the cluster scales.
Without memoization, search time rises to 12.3~s on 128~GPUs; with memoization it is capped at 1.9~s ($\sim$6.5$\times$ reduction), ensuring that re-planning remains a tiny fraction of the overall iteration time.

\begin{table}[t]
    \centering
    \caption{Worker state transition breakdown in the Elastic Hybrid Pool.}
    \label{tab:worker_transition}
    \scriptsize
    \setlength{\tabcolsep}{2pt}
    \renewcommand\cellgape{\Gape[1.5pt]}
    \begin{tabular}{@{}l l c l@{}}
        \toprule
        \textbf{Direction} & \textbf{Operation} & \textbf{Overhead}  & \textbf{Mechanism} \\
        \midrule
        \multirow{1}{*}{\makecell{Training\\$\rightarrow$\\Rollout}}
         & Training context teardown & 15 ms & \makecell[l]{Release local resources} \\
         & vLLM instance activation & 10.8 s & \makecell[l]{Enable CUDA graph replay;\\Reuse torch.compile cache} \\
        \midrule
        \multirow{4}{*}{\makecell{Rollout\\$\rightarrow$\\Training}}
         & Snapshot capture & 3.6 s  & Per-step, overlapped with training \\
         & State reload & 4.4 s & Proceeds via RDMA\\
         & Gradient channel setup & $\lesssim$20 ms & One-time registration \\
         & Training state alignment & 20 ms & Join via zero-gradient sync \\
        \bottomrule
    \end{tabular}
    \vspace{2pt}
    \raggedright
\end{table}

\subsection{Cost Evaluator Fidelity}
\begin{figure}[t]
    \centering
    \includegraphics[width=1.02\columnwidth]{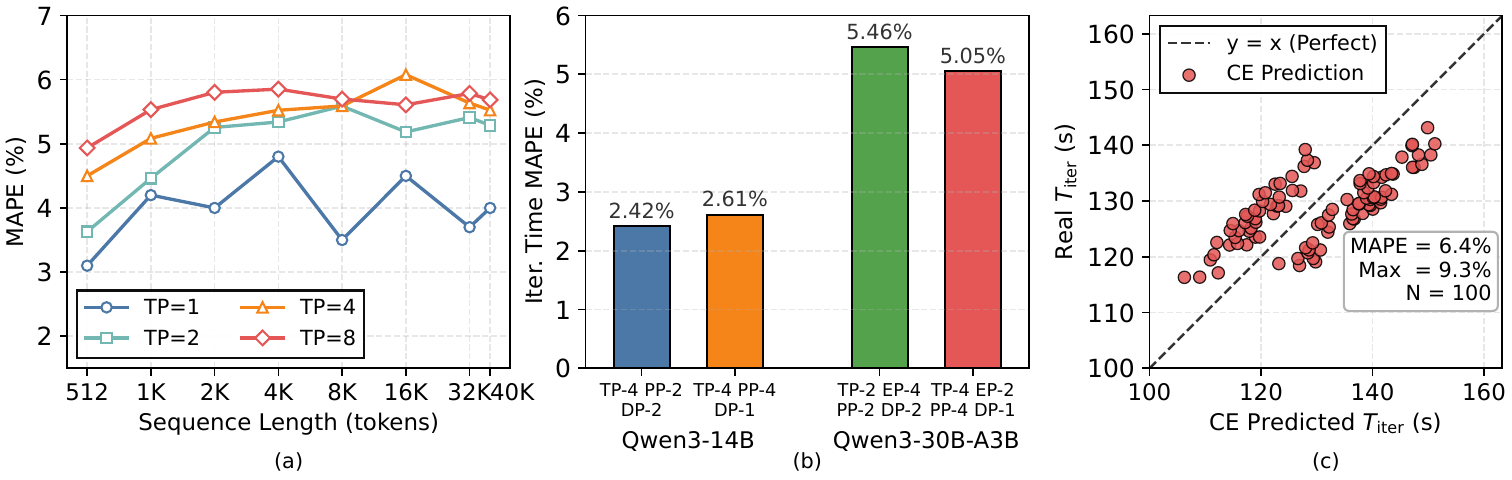}
    \caption{Cost Evaluator fidelity validation.
    (a)~Rollout step-time prediction accuracy (MAPE).
    (b)~Training iteration-time MAPE.
    (c)~End-to-end scatter plot of CE-predicted versus measured iteration time over 100 estimations.}
    \label{fig:ce_validation}
\end{figure}

Figure~\ref{fig:ce_validation} assesses the prediction fidelity of \sys's Cost Evaluator (CE).
In panel~(a), rollout step-time MAPE remains between 3.1\% and 5.9\% across sequence lengths and TP degrees.
In panel~(b), training iteration-time MAPE stays at 2.4--2.6\% on Qwen3-14B and 5.0--5.5\% on Qwen3-30B-A3B under heterogeneous micro-batch loads.
In panel~(c), CE predictions closely track 100 real iteration times with 6.35\% average MAPE and 9.30\% maximum error.

\section{Related Work}
\label{sec:related}

Table~\ref{tab:related_comparison} summarizes the comparison between \sys and representative related work.

\noindent \textbf{RL frameworks for LLM post-training.}
Colocated frameworks~\cite{verl, openrlhf, slime,deepspeed} alternate training and rollout on shared GPUs, while disaggregated frameworks~\cite{areal, Laminar, streamrl,RollArt, ROSE, Distrl,asyncflow} isolate stages onto dedicated clusters to execute asynchronously. 
Concurrent work DynaRL~\cite{DynaRL} dynamically moves GPUs across stages, but confines elasticity to 1D GPU counts without adapting internal parallelism strategies. 
\sys instead periodically re-plans both the cross-stage GPU allocation and each cluster's internal parallelism strategy, and elastically rebalances workers through a non-blocking protocol.

\noindent \textbf{Rollout scheduling and long-tail mitigation.}
To mitigate long-tails, Heddle~\cite{Heddle} and streamrl~\cite{streamrl} employ auxiliary models for length prediction, which suffer representation drift under policy updates. 
Seer~\cite{Seer} and RhymeRL~\cite{RhymeRL} rely on intra-group or cross-epoch response similarity, which fails when mid-trajectory tool returns heavily alter sequence length. 
\sys instead exploits runtime tool outcomes for causality-guided bucket routing.

\noindent \textbf{Distributed parallelization and planning.}
Distributed pre-training managers~\cite{alpa,galvatron,sailor,lyra} optimize parallelism strategies for static workloads. 
However, they target single-phase training and do not consider the dynamic, coupled execution of RL rollout and training stages.

\begin{table}[t]
\centering
\caption{Comparison with representative RL post-training and inference scheduling frameworks. }
\label{tab:related_comparison}
\small
\resizebox{\columnwidth}{!}{%
\begin{tabular}{l cccccc}
\toprule
\textbf{} & \textbf{\sys} & \textbf{\makecell{DynaRL\\\cite{DynaRL}}} & \textbf{\makecell{AReal\\\cite{areal}}} & \textbf{\makecell{Laminar\\\cite{Laminar}}} & \textbf{\makecell{Heddle\\\cite{Heddle}}} & \textbf{\makecell{veRL\\\cite{verl}}} \\
\midrule
\rowcolor{gray!15}
\multicolumn{7}{l}{\textbf{Cross-Stage Coordination}} \\
Disaggregated Execution             & \checkmark & \checkmark & \checkmark & \checkmark & $-$   & $-$   \\
Cross-Stage Resource Planning     & \checkmark & \checkmark & $\times$   & $\times$   & $\times$   & $\times$   \\
Multi-Dim Parallelism Planning   & \checkmark & $\times$   & $\times$   & $\times$   & $\times$   & $\times$   \\
Non-Blocking Worker Migration       & \checkmark & $\times$   & $\times$   & $\times$   & $\times$   & $\times$   \\
\rowcolor{gray!15}
\midrule
\multicolumn{7}{l}{\textbf{Rollout Parallelism \& Scheduling}} \\
Length-Aware Scheduling             & \checkmark & $\times$   & $\times$ & $\times$   & $-$   & $\times$ \\
Heterogeneous Parallelism           & \checkmark & $\times$   & $\times$   & $\times$   & \checkmark   & $\times$   \\
\bottomrule
\end{tabular}%
}
\end{table}

\section{Conclusion}

This paper presents \sys, an adaptive runtime for agentic RL post-training.
\sys introduces a Causality-Guided Bucket Scheduler that routes requests across heterogeneous rollout buckets based on  causal signals from tool-returns.
Our evaluation on both GPU and NPU clusters demonstrates that \sys delivers substantial throughput gains and accelerated reward convergence. \sys has been open sourced.

\bibliographystyle{ACM-Reference-Format}
\bibliography{main}
\appendix

\section{Rationale of Rollout Cluster Parallelism Design}
\label{appendix:pp_ep}

In \sys, the rollout cluster is designed with heterogeneous Tensor Parallelism (TP) as the primary parallelism strategy, while Pipeline Parallelism (PP) and Expert Parallelism (EP) are not explicitly included in the current rollout configuration.
This appendix justifies this design choice from three perspectives: workload characteristics, problem orthogonality, and extensibility.

\subsection{Workload Characteristics of Agentic RL Rollout}

Agentic RL rollout is dominated by autoregressive decoding of long, variable-length trajectories. The key performance bottlenecks are (1)~memory bandwidth and KV-cache pressure, especially for long sequences~\cite{distserve, Sarathi-Serve}, and (2)~straggler effects caused by a small fraction of long-tailed trajectories~\cite{Laminar,Seer,streamrl,Heddle,stage_fusion,roll_packer}.
TP partitions model weights and the KV cache across GPUs, directly alleviating memory bottlenecks and balancing computation~\cite{TP-tradeoff}. 
In contrast, PP splits model layers across devices, introducing sequential dependencies that do not help with stragglers and often amplify them: a single long request holding a pipeline stage blocks all downstream stages until completion~\cite{amd2025vllmmoe}.

\subsection{Problem Orthogonality: Why TP Suffices for Long-Tailed Mitigation}

\sys's core contribution is mitigating straggler effects in heterogeneous rollout. 
As shown in Figure~\ref{fig:heterogeneity}, TP size directly determines the throughput trade-off: small TP (e.g., 1, 2) incurs low communication overhead and is ideal for short sequences, whereas large TP (e.g., 4, 8) offers better memory scalability and sustained throughput for long sequences.

PP does not offer this trade-off. 
It distributes layers sequentially, so a single long request still incurs high latency through all stages, and pipeline bubbles worsen under length imbalance. 
EP, while useful for Mixture-of-Experts (MoE) models, addresses expert load balancing rather than sequence-length-induced stragglers. 
Thus, PP and EP are orthogonal to \sys's primary optimization target.

\subsection{Extensibility: How \sys Can Support PP and EP}
\sys's heterogeneous rollout cluster design is not inherently closed to PP or EP.
Extending the rollout cluster to support PP and EP follows the following principles.

\paragraph{Pipeline Parallelism in rollout.} If a model is too large to fit within a single node, PP can be introduced inside each heterogeneous bucket as an additional intra-bucket dimension. 
The bucket then consumes $tp \times pp$ GPUs rather than $tp$ GPUs. 
Extending the rollout DP (Algorithm~\ref{alg:hierarchical_planner}) to support this requires two localized changes: (1)~expanding the candidate set $\mathcal{T}$ to enumerate feasible $(TP, PP)$ combinations with their corresponding total GPU counts, 
and (2)~updating the CE interface to accept a PP argument so that the latency model accounts for inter-stage pipeline bubbles. Please note that vidur's latency model natively supports PP.
The DP's optimal-substructure property remains unchanged because the state transition still consumes a fixed number of GPUs per bucket. 
The routing of Causality-Guided Bucket Scheduler can also be adapted to this setting, as the heterogeneous buckets retain their relative affinity for long and short sequences.

\paragraph{Expert Parallelism in rollout.} 
For MoE models, EP can be transparent to \sys's planner because the inference engine handles it automatically. 
In vLLM,  the engine automatically enables EP when the number of experts exceeds the TP size and is divisible by it, 
and to fall back to TP-only execution otherwise to avoid unnecessary all-to-all communication. 
Because \sys's heterogeneous rollout already traverses a range of TP configurations (e.g., TP=1, 2, 4, 8), each bucket implicitly covers both the EP-enabled and EP-disabled regimes: 
a smaller TP may trigger EP, while a larger TP may not. The planner therefore does not need to enumerate EP as an explicit dimension; extending rollout to MoE models amounts to letting the inference engine select EP automatically within each bucket. The inter-bucket scheduling mechanism remains unchanged.

\section{Offline Profile and Prefix-Tree Lifecycle}
\label{appendix:profile}

This appendix details the offline profile phase that precedes formal training, and provides empirical measurements on prefix-tree coverage and fallback frequency that complement the main text.

\subsection{Offline Profile Phase}
\label{appendix:subsec:profile}

Before the first training iteration, \sys executes an \textit{offline profile phase} whose primary goal is to obtain the initial sequence length distribution of the target workload.
Using the initial policy model, \sys generates trajectories for the entire training corpus; the resulting length distribution then drives two downstream tasks: (1)~calibrating the cost evaluator (CE) and deriving the initial resource allocation plan, and (2)~bootstrapping the prefix tree with historical trajectory data.

\paragraph{CE calibration.}
The profile trajectories are fed to the CE to fit the per-operator polynomial latency models described in Section~\ref{section:CE}.
The CE records the execution time of each layer type (Linear, Attention, etc.) at varying sequence lengths, yielding the coefficients $\gamma, \eta, \theta$ for the polynomial $T(L)$.
These models then drive the planner's initial DP search (Algorithm~\ref{alg:hierarchical_planner}), producing the first rollout partition $P_{\text{rollout}}$ and training strategy $S_{\text{train}}$.

\paragraph{Prefix-tree bootstrap.}
Every profile trajectory is inserted into the prefix tree following the procedure described in Phase~3 of Section~\ref{CGBS}.
Consequently, every prompt in the training corpus is present in the tree from the very first training step.
The profile phase produces $N \times K$ trajectories in total.
These trajectories are inserted into the trie, which naturally shares prefixes: any two trajectories that follow the same tool-return sequence up to depth $d$ share the path from the root to depth $d$.
Consequently, the trie compactly represents the full history without materializing $N \times K$ independent branches.
Table~\ref{tab:profile_stats} shows the resulting tree statistics for the three benchmarks.

\begin{table}[h]
\centering
\caption{Offline profile statistics and initial prefix-tree coverage.}
\label{tab:profile_stats}
\small
\begin{tabular}{@{}l r r r@{}}
\toprule
\makecell{Benchmark} & \makecell{\#Prompts} & \makecell{Profile Time} & \makecell{Avg.~Tree Depth} \\
\midrule
\makecell{Search-R1} & \makecell{17,000} & \makecell{$\sim$1.24~h} & \makecell{5.2} \\
\makecell{R2E-Gym} & \makecell{8,135} & \makecell{$\sim$3.8~h} & \makecell{22.0} \\
\makecell{DAPO-Math-17K} & \makecell{17,000} & \makecell{$\sim$2.8~h} & \makecell{2.5} \\
\bottomrule
\end{tabular}
\end{table}

All profile experiments use Qwen3-14B on NVIDIA A800 80GB GPUs, deployed on 48 GPUs with TP-8 and DP-6, processing a batch of 1,024 prompts.
\paragraph{Reusability and Break-Even Analysis.}                                                    
    Following standard methodology in distributed ML systems~\cite{alpa,galvatron,vidur}, the end-to-end
  training and convergence times reported in Section~\ref{evaluation} evaluate active online training   
  execution and do not include this upfront profiling phase.                                            
    This exclusion is justified by the high reusability of the offline profile: because the profile     
  depends strictly on the initial model weights and prompt corpus, a single profile is generated once   
  and reused across all subsequent hyperparameter tuning runs (e.g., learning rates, KL penalties,      
  reward scaling), algorithmic variations (e.g., GRPO vs.~PPO), and ablation experiments on that        
  benchmark.                                                                                            
                                                                                                        
    Furthermore, even under the worst-case scenario where a profile is consumed by only a               
  \textit{single} training run, \sys achieves break-even rapidly due to its substantial end-to-end      
  throughput advantages:                                                                                
    \begin{itemize}[nosep,leftmargin=*]                                                                 
        \item On \textit{Search-R1}, \sys saves 7.3~h of online training time compared to the strongest 
  dynamic baseline DynaRL (16.2~h vs.~23.5~h). Factoring in the 1.24~h profiling cost, \sys achieves a  
  net savings of 6.06~h (a $1.35\times$ speedup), reaching the break-even point in under 20\% of the    
  training run.                                                                                         
        \item On \textit{R2E-Gym}, \sys cuts training time by 19.8~h over DynaRL (54.0~h vs.~73.8~h),   
  which easily amortizes the 3.8~h profile overhead and yields a net saving of 16.0~h ($1.28\times$ end-
  to-end speedup).                                                                                      
    \end{itemize}                                                                                       
    Once online training commences, \sys adapts to policy drift through sub-second runtime trie updates 
  ($<$1~s, \S\ref{appendix:subsec:update}), completely eliminating the need to re-execute offline       
  profiling during the training lifecycle.

\subsection{Tree Update and Policy-Drift Adaptation}
\label{appendix:subsec:update}

As RL training progresses, the policy model evolves and the trajectory distribution shifts.
\sys updates the tree by inserting each completed trajectory along its return-state sequence (Phase~3 of Section~\ref{CGBS}), 
creating child nodes for unseen states as needed and recording the remaining length at every visited node.
Consequently, subsequent lookups immediately benefit from the latest observations.
To further adapt to policy drift and amortize stale statistics, \sys additionally rebuilds the tree offline at regular intervals (every $T$ training steps, default $T=50$) using the most recent trajectories.
Because the tree is a lightweight lookup structure (a trie with scalar statistics per node), both incremental insertion and full rebuild take less than one second and incur no online overhead.
This dual update mechanism is the primary means through which Causality-Guided Bucket Scheduler adapts to policy drift without retraining a parametric prediction model.

\subsection{Fallback Frequency and Tree Coverage}
\label{appendix:subsec:fallback}

A runtime routing lookup walks the tree from the root along the tool-return sequence observed so far.
If the next return state has never been seen under the current prefix, the scheduler falls back to the parent node's statistics (Section~\ref{CGBS}).
We instrumented \sys on the Search-R1 workload to measure how often this fallback occurs.

\begin{table}[h]
\centering
\caption{Fallback frequency per training phase (Search-R1).}
\label{tab:fallback}
\small
\begin{tabular}{@{}l r r r@{}}
\toprule
\makecell{Training\\Phase} & \makecell{Avg.\\~Tree Lookups} & \makecell{Avg.\\~Fallbacks} & \makecell{Fallback\\Ratio} \\
\midrule
\makecell{Steps 1--50} & \makecell{3,952} & \makecell{236} & \makecell{5.98\%} \\
\makecell{Steps 51--100} & \makecell{4,248} & \makecell{174} & \makecell{4.10\%} \\
\makecell{Steps 101--200} & \makecell{9,424} & \makecell{347} & \makecell{3.68\%} \\
\makecell{Steps 201--400} & \makecell{16,492} & \makecell{462} & \makecell{2.80\%} \\
\bottomrule
\end{tabular}
\end{table}

Table~\ref{tab:fallback} shows that the fallback ratio decreases as training proceeds.
During the initial phase (steps 1--50), the tree contains only the offline profile data, yet the fallback ratio is below 6\%.
As the policy explores new tool-return combinations, the tree is periodically rebuilt with fresh trajectories, and the fallback ratio drops to under 4\% by the later stages of training.
This empirically confirms that the offline profile provides sufficient initial coverage, and that periodic updates effectively amortize the sparse-revisit concern.

\section{Decoupled Communication Domain}
\label{appendix:decoupled_comm}

This appendix details the implementation of \sys's decoupled communication domain, explains why membership changes in the inter-replica plane do not block the core training pool.

\subsection{RDMA-based All-Reduce Protocol Implementation}
\sys realizes the inter-replica gradient exchange through a RDMA-based All-Reduce protocol inspired by FT-HSDP~\cite{FT-HSDP}.
It departs from NCCL's static communicator model by adopting a hybrid control-plane/data-plane architecture.

\paragraph{Control plane (CPU async thread).}
The CPU thread manages all membership-related complexity.
It maintains the active replica set through a consensus service, detects failures via heartbeat timeouts, and dynamically rewires the ring topology by updating per-rank RDMA send/receive buffer descriptors.
Because these operations are lightweight metadata updates, they complete in milliseconds without touching GPU state.

\paragraph{Data plane (GPU stream).}
The GPU stream executes only data copy and reduction kernels using a ring algorithm (ReduceScatter followed by AllGather).
To prevent GPU hangs, the kernel busy-polls a host-pinned ready flag with a software timeout rather than blocking indefinitely.
When the flag signals data arrival, the kernel performs in-GPU reduction and forwards the result to the next neighbor.

\subsection{Why the Core Pool Remains Unblocked}
The decoupled design guarantees that the intra-replica NCCL collective never stalls for three reasons.

\textit{Reason 1: metadata-only membership updates.}
When a replica exits the core training pool, the CPU removes it from the inter-replica member list and recomputes the ring order.
Core training workers are completely unaware of this change; the CPU merely refreshes neighbor-address tables in host-pinned descriptors before the next gradient-exchange launch.
Unlike NCCL, which requires destroying and recreating the entire communicator (a multi-second operation), \sys's update is a microsecond-scale metadata edit.

\textit{Reason 2: ring algorithm tolerates absent nodes.}
The ring AllReduce algorithm does not require all participants to be simultaneously online.
Once the exiting replica is excluded from the neighbor map, the data plane naturally skips it: each core training worker sends to and receives from its live predecessor and successor only.
Because the GPU kernel derives its peer list from the CPU-updated metadata at launch time, it never issues RDMA operations to a dead endpoint and therefore never blocks on a timeout.

\textit{Reason 3: non-blocking catch-up.}
When a replica is ready to join the training process, the CPU inserts it into the ring at an arbitrary position and updates only its two immediate neighbors' routing tables.
Core training workers refresh their RDMA address mappings at the next gradient-exchange boundary; this refresh is a single memory write visible to the GPU kernel within microseconds.

\section{Non-blocking Joining}
\label{appendix:rejoin}

This appendix formalizes the aggregation rule of the non-blocking joining protocol 
and proves that it preserves mathematical equivalence across both active elastic training and joining windows. 
We further analyze optimizer-state consistency under asynchronous recovery and provide empirical validation.

\subsection{Aggregation Rule and Optimization Invariance}

\noindent \textbf{Workload model and fixed global batching.}
In agentic RL post-training (e.g., GRPO/PPO), each training step consumes a fixed global batch of rollout trajectories $\mathcal{B}$, comprising $S = |\mathcal{B}|$ samples (e.g., 4096 samples). The target optimization gradient is the exact sample-averaged gradient across this fixed batch:
\begin{equation}
\bar{g}^* = \frac{1}{S} \sum_{s \in \mathcal{B}} \nabla \ell(s; \theta).
\end{equation}
When the training cluster operates with $N_c$ core data-parallel replicas and $N_h$ active hybrid replicas (total $N = N_c + N_h$ replicas), \sys partitions $\mathcal{B}$ disjointly across all live replicas:
\begin{equation}
\mathcal{B} = \left( \bigcup_{i=1}^{N_c} \mathcal{B}_i^{\text{core}} \right) \cup \left( \bigcup_{h=1}^{N_h} \mathcal{B}_h^{\text{hyb}} \right),
\end{equation}
where $|\mathcal{B}_i^{\text{core}}| = S_i^{\text{core}}$ and $|\mathcal{B}_h^{\text{hyb}}| = S_h^{\text{hyb}}$ such that $\sum_{i=1}^{N_c} S_i^{\text{core}} + \sum_{h=1}^{N_h} S_h^{\text{hyb}} = S$. Adding hybrid DP replicas increases DP parallelism to accelerate the step by reducing each replica's local micro-batch count, without expanding the global batch size $S$.

\noindent \textbf{Topology mapping and two-tier gradient aggregation.}
Each active hybrid DP replica $h \in \{1, \dots, N_h\}$ is paired to a core DP replica $\pi(h) \in \{1, \dots, N_c\}$. The control plane distributes hybrid replicas evenly across core replicas such that $|\mathcal{H}_i| \le \lceil N_h / N_c \rceil$, where $\mathcal{H}_i = \{h \mid \pi(h) = i\}$. 
Because each DP replica consists of $R_{\text{train}} = TP \times PP$ workers, this replica pairing $\pi$ is physically established by configuring peer-to-peer RDMA side channels between corresponding workers (i.e., workers with matching intra-replica TP/PP positions) in hybrid replica $h$ and core replica $\pi(h)$.
Each replica $k$ computes the unnormalized gradient sum over its assigned samples:
\begin{equation}
G_k = \sum_{s \in \mathcal{B}_k} \nabla \ell(s; \theta).
\end{equation}
Hybrid DP replica $h$ routes its gradient sum $G_h^{\text{hyb}}$ to core DP replica $\pi(h)$ through the RDMA side channel. Core DP replica $i$ accumulates external gradients into its local sum:
\begin{equation}
\label{eq:local_accum}
\tilde{G}_i = G_i^{\text{core}} + \sum_{h \in \mathcal{H}_i} G_h^{\text{hyb}}.
\end{equation}
The core pool then executes an intra-core All-Reduce (Sum):
\begin{equation}
\label{eq:global_sum}
G_{\text{total}} = \sum_{i=1}^{N_c} \tilde{G}_i = \sum_{i=1}^{N_c} G_i^{\text{core}} + \sum_{h=1}^{N_h} G_h^{\text{hyb}} = \sum_{s \in \mathcal{B}} \nabla \ell(s; \theta).
\end{equation}
The effective gradient for parameter updates is normalized by the fixed global sample count:
\begin{equation}
\label{eq:global_norm}
\bar{g} = \frac{G_{\text{total}}}{S}.
\end{equation}

\noindent \textbf{Mathematical equivalence across elastic states.}
This sample-normalized sum aggregation guarantees exact mathematical equivalence ($\bar{g} \equiv \bar{g}^*$) under both operational phases:
\begin{itemize}[leftmargin=*,nosep]
\item \textit{Active Elastic Training ($G_h \neq 0$):} From Eq.~\eqref{eq:global_sum} and \eqref{eq:global_norm}, $\bar{g}$ is strictly identical to $\bar{g}^*$. Because the normalization denominator $S$ remains fixed, the gradient magnitude, effective sample weights, and loss landscape are strictly invariant to $N_h$. Hence, elastic transitions require no learning-rate rescaling or warmup adjustments.
\item \textit{Non-blocking Joining Window ($G_h = 0$):} During state restoration, a joining DP replica has not completed loading its snapshot, so it is assigned zero samples ($S_h^{\text{hyb}} = 0$) and emits a zero placeholder ($G_h^{\text{hyb}} = 0$). The global batch $\mathcal{B}$ is partitioned among the remaining live replicas (i.e., the $N_c$ core replicas). In Eq.~\eqref{eq:local_accum}, $\tilde{G}_i = G_i^{\text{core}} + 0$. The All-Reduce sums to the exact total gradient over all $S$ samples, and division by $S$ yields the identical unbiased gradient $\bar{g}^*$.
\end{itemize}

\subsection{Two-Phase Version Alignment and Optimizer Consistency}
\label{appendix:subsec:state_alignment}

To eliminate state staleness, race conditions, or optimizer moment divergence when a hybrid DP replica joins without stalling ongoing training, \sys employs a deterministic two-phase version alignment protocol.

\paragraph{Phase 1: Intra-step snapshot reload.}
As quantified in Table~\ref{tab:worker_transition}, reloading the complete state snapshot $(\theta_t, m_t, v_t)$ from CPU host memory via RDMA requires only 4.4~s (executed concurrently across all constituent workers of the replica), which corresponds to  $\sim$3.8\% of an active training step (average step time $\sim$116.6~s).
The joining replica initiates this reload at the onset of step $t$. 
By second 4.4, the replica reconstructs the exact model parameters $\theta_t$ and optimizer moments $(m_t, v_t)$ (sharded according to the model-parallel configuration across its constituent workers). 
Crucially, throughout the remaining $\sim$112~s of step $t$, the core replicas are engaged in forward and backward computation on their local micro-batches; their physical parameters and optimizer states remain strictly frozen at version $t$. 
Thus, the joining replica's restored state is perfectly concurrent with the core pool throughout step $t$, completely avoiding stale-parameter issues.

\paragraph{Phase 2: Lock-step step transition.}
Because the joining replica was loading state during the first 4.4~s of step $t$, it does not participate in step $t$'s forward pass ($S_h = 0$) and emits a zero placeholder $G_h = 0$ via the side channel. 
When the core pool completes step $t$'s All-Reduce, it computes the nominal global gradient $\bar{g}_t$ (Eq.~\ref{eq:global_norm}) and forwards $\bar{g}_t$ to the joining replica through the side channel. 
Both the core replicas and the joining replica then execute the identical optimizer update on identical inputs $(\theta_t, m_t, v_t, \bar{g}_t)$. 
Under Adam, both sides compute:
\begin{align}
m_{t+1} &\leftarrow \beta_1 m_t + (1-\beta_1)\bar{g}_t, \\
v_{t+1} &\leftarrow \beta_2 v_t + (1-\beta_2)(\bar{g}_t)^2, \\
\theta_{t+1} &\leftarrow \theta_t - \eta \cdot \frac{m_{t+1}}{\sqrt{v_{t+1}} + \epsilon}.
\end{align}
Under plain SGD, this update simplifies to $\theta_{t+1} \leftarrow \theta_t - \eta \bar{g}_t$. 
Because the deterministic update functions and inputs are identical, parameters $\theta_{t+1}$ and optimizer moments $(m_{t+1}, v_{t+1})$ advance in lock-step to step $t+1$ without requiring full-state re-broadcasting.
At the boundary of step $t+1$, the joining replica's state is mathematically identical to the core pool. 
From step $t+1$ onward, the replica operates as an active DP replica, computing non-zero gradients on its assigned batch partition.

\begin{figure}[t]
\centering
\includegraphics[width=\columnwidth]{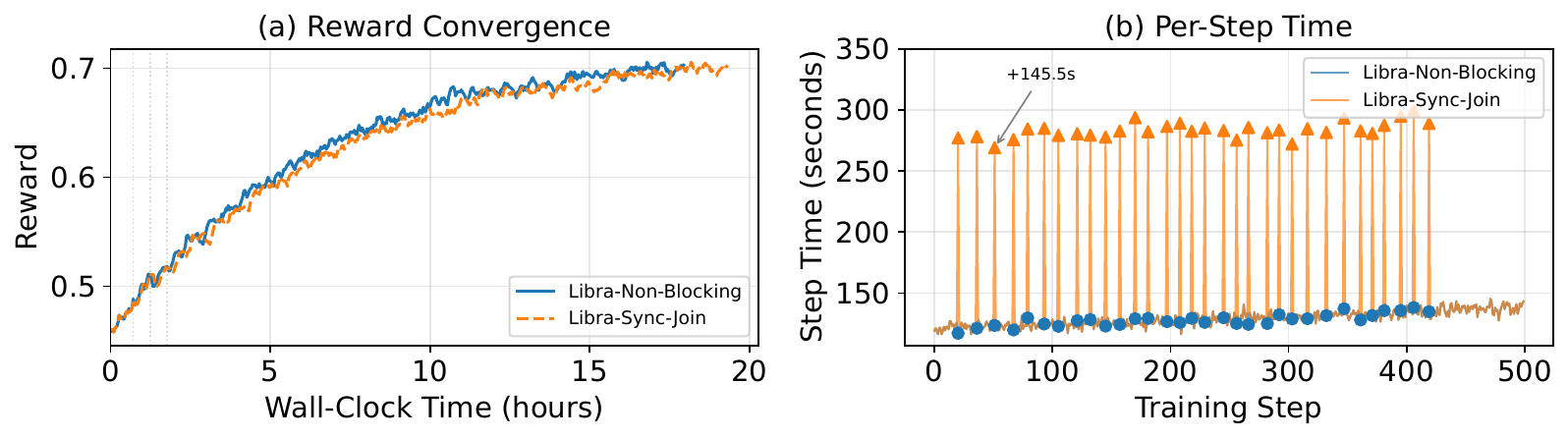}
\caption{Validation of non-blocking joining on Search-R1.
(a)~Reward convergence over wall-clock time: \sys-Non-Blocking reaches the same reward faster than \sys-Sync-Join.
(b)~Per-step time: \sys-Sync-Join incurs a $\sim$150~s spike at each transition step, while \sys-Non-Blocking remains flat.}
\label{fig:joining_validation}
\end{figure}

\subsection{Empirical Validation}

To empirically validate that non-blocking joining does not degrade training convergence or parameter accuracy, 
we run a controlled paired experiment on the Search-R1 workload (Qwen3-14B, 500 training steps).
Both configurations use identical planner decisions, heterogeneous TP buckets, and Causality-Guided Bucket Scheduler routing; 
the only difference is how worker transitions are handled.

\noindent \textbf{\sys-Non-Blocking.} 
The core training cluster continues advancing while joining replicas reload their snapshots asynchronously via the zero-gradient protocol described above.

\noindent \textbf{\sys-Sync-Join.} 
To emulate the behavior of conventional training frameworks (e.g., FSDP~\cite{fsdp,fsdp2}, Megatron-LM~\cite{megatron}),
whenever a hybrid DP replica transitions back to training, the entire core training cluster is torn down and rebuilt with the new replica included. 
This involves destroying the existing NCCL communicators, re-initializing the distributed training context, redistributing model and optimizer states, and re-warming the first training step. 

\noindent \textbf{Convergence and numerical equivalence.} 
Figure~\ref{fig:joining_validation}(a) shows that \sys-Non-Blocking and \sys-Sync-Join reach the same final reward ($0.70$) after 500 training steps. 
To strictly verify mathematical state equivalence beyond reward convergence, we tracked the maximum absolute parameter difference $\max |\theta_{\text{hyb}} - \theta_{\text{core}}|$ and optimizer moment differences across all model layers immediately following each joining transition. The difference remains below $1.2 \times 10^{-7}$ (within FP32 numerical precision), confirming that non-blocking joining achieves exact state alignment without parameter drift or gradient distortion.

\noindent \textbf{Overhead quantification.} 
Figure~\ref{fig:joining_validation}(b) shows that \sys-Sync-Join incurs a $\sim$150~s spike at every transition step, 
reflecting the full cost of global communicator rebuild and state redistribution.

\end{document}